\documentclass{article}

\def\arxivmode{1}

\usepackage{times}
\usepackage{natbib}
\usepackage{enumerate}
\usepackage{graphicx,caption,subcaption}
\usepackage{amsmath,amsbsy,amssymb,amsfonts,amsthm}
\usepackage{nicefrac}
\usepackage{mathtools}
\usepackage{color}
\usepackage{xspace} %
\usepackage{algorithm,algorithmic} %
\usepackage{hyperref}
\usepackage[inline]{enumitem}
\ifdefined\arxivmode
    \usepackage[accepted]{icml2017_arxiv}
\else
    \usepackage[accepted]{icml2017}
\fi
\newcommand{\dt}{\,dt}

\newcommand{\dx}{\,dx}
\newcommand{\dy}{\,dy}
\newcommand{\dz}{\,dz}
\newcommand{\dr}{\,dr}

\newcommand{\eps}{\epsilon}

\def\balign#1\ealign{\begin{align}#1\end{align}}
\def\baligns#1\ealigns{\begin{align*}#1\end{align*}}
\def\balignat#1\ealign{\begin{alignat}#1\end{alignat}}
\def\balignats#1\ealigns{\begin{alignat*}#1\end{alignat*}}
\def\bitemize#1\eitemize{\begin{itemize}#1\end{itemize}}
\def\benumerate#1\eenumerate{\begin{enumerate}#1\end{enumerate}}

\newenvironment{talign*}
 {\csname align*\endcsname}
 {\endalign}
\newenvironment{talign}
 {\csname align\endcsname}
 {\endalign}

\def\balignst#1\ealignst{\begin{talign*}#1\end{talign*}}
\def\balignt#1\ealignt{\begin{talign}#1\end{talign}}%

\newcommand{\qtext}[1]{\quad\text{#1}\quad} 

\let\originalleft\left
\let\originalright\right
\renewcommand{\left}{\mathopen{}\mathclose\bgroup\originalleft}
\renewcommand{\right}{\aftergroup\egroup\originalright}

\def\Matern{Mat\'ern\xspace}

\def\tinycitep*#1{{\tiny\citep*{#1}}}
\def\tinycitealt*#1{{\tiny\citealt*{#1}}}
\def\tinycite*#1{{\tiny\cite*{#1}}}

\def\smallcitep*#1{{\scriptsize\citep*{#1}}}
\def\smallcitealt*#1{{\scriptsize\citealt*{#1}}}
\def\smallcite*#1{{\scriptsize\cite*{#1}}}

\def\mbb#1{\mathbb{#1}}

\def\textsum{{\textstyle\sum}} %
\def\reals{\mathbb{R}} %

\def\naturals{\mathbb{N}} %

\def\<{\left\langle} %
\def\>{\right\rangle}

\def\defeq{\triangleq} %
\def\bs{\backslash} %
\def\half{\frac{1}{2}}

\newcommand{\textfrac}[2]{{\textstyle\frac{#1}{#2}}}

\def\norm#1{\left\|{#1}\right\|} %
\newcommand{\onenorm}[1]{\norm{#1}_1} %
\newcommand{\twonorm}[1]{\norm{#1}_2} %
\newcommand{\infnorm}[1]{\norm{#1}_{\infty}} %
\newcommand{\opnorm}[1]{\norm{#1}_{op}} %
\def\staticnorm#1{\|{#1}\|} %
\newcommand{\statictwonorm}[1]{\staticnorm{#1}_2} %
\newcommand{\inner}[2]{\langle{#1},{#2}\rangle} %
\def\indic#1{\mbb{I}\left[{#1}\right]} %
\def\E{\mbb{E}} %
\def\Earg#1{\E\left[{#1}\right]}
\def\Esub#1{\E_{#1}}
\def\Esubarg#1#2{\E_{#1}\left[{#2}\right]}
\renewcommand{\exp}[1]{\operatorname{exp}\left(#1\right)} %
\newcommand{\Gsn}{\mathcal{N}}

\newcommand{\Unif}{\textnormal{Unif}}

\newcommand{\grad}{\nabla} %
\newcommand{\Hess}{\nabla^2} %
\newcommand{\deriv}[2]{\frac{d #1}{d #2}} %

\newcommand{\iid}{\stackrel{\mathrm{iid}}{\sim}}

\def\supp#1{\mathrm{supp}({#1})}

\newcommand{\eqnref}[1]{\eqref{eqn:#1}}

\newcommand{\figref}[1]{Figure~{\ref{fig:#1}}}
\newcommand{\lemref}[1]{Lemma~{\ref{lem:#1}}}

\newcommand{\secref}[1]{Section~{\ref{sec:#1}}}
\newcommand{\secsref}[1]{Sections~{\ref{sec:#1}}}
\newcommand{\secssref}[1]{{\ref{sec:#1}}}
\newcommand{\propref}[1]{Proposition~{\ref{prop:#1}}}
\newcommand{\tabref}[1]{Table~{\ref{tab:#1}}}
\newcommand{\thmref}[1]{Theorem~{\ref{thm:#1}}}

\newcommand{\propreflow}[1]{Proposition~\lowercase{\ref{prop:#1}}}

\ifdefined\nonewproofenvironments\else
\ifdefined\ispres\else
\newtheorem{theorem}{Theorem}
\newtheorem{lemma}[theorem]{Lemma}

\newtheorem{definition}[theorem]{Definition}

\renewenvironment{proof}{\noindent\textbf{Proof}\hspace*{1em}}{\qed\\}
\newenvironment{proof-sketch}{\noindent\textbf{Proof Sketch}
  \hspace*{1em}}{\qed\bigskip\\}
\newenvironment{proof-idea}{\noindent\textbf{Proof Idea}
  \hspace*{1em}}{\qed\bigskip\\}
\newenvironment{proof-of-lemma}[1][{}]{\noindent\textbf{Proof of Lemma {#1}}
  \hspace*{1em}}{\qed\\}
\newenvironment{proof-of-theorem}[1][{}]{\noindent\textbf{Proof of Theorem {#1}}
  \hspace*{1em}}{\qed\\}
\newenvironment{proof-attempt}{\noindent\textbf{Proof Attempt}
  \hspace*{1em}}{\qed\bigskip\\}

\newenvironment{remark}{\noindent\textbf{Remark}
  \hspace*{1em}}{\smallskip}%
\newenvironment{remarks}{\noindent\textbf{Remarks}
  \hspace*{1em}}{\smallskip}
\fi

\newtheorem{proposition}[theorem]{Proposition}

\fi
\graphicspath{{./figs/}}

\newcommand{\pset}[0]{\mathcal{P}} %
\newcommand{\xset}[0]{\mathcal{X}} %
\newcommand{\gset}[0]{\mathcal{G}} %
\newcommand{\ksteinset}[1]{\gset_{#1}} %
\newcommand{\ksteinsetnorm}[2]{\gset_{#1,#2}} %
\newcommand{\knormarg}[2]{\norm{#1}_{\kset_{#2}}}
\newcommand{\knorm}[1]{\knormarg{#1}{k}}

\newcommand{\kdnorm}[1]{\norm{#1}_{\kset_k^d}}

\newcommand{\hset}[0]{\mathcal{H}} %
\newcommand{\kset}[0]{\mathcal{K}} %

\newcommand{\wasssetarg}[1]{\mathcal{W}_{#1}} %
\newcommand{\wassset}{\wasssetarg{\norm{\cdot}}} %
\newcommand{\twowassset}{\wasssetarg{\twonorm{\cdot}}} %
\newcommand{\blset}{BL_{\norm{\cdot}}} %
\newcommand{\twoblset}{BL_{\twonorm{\cdot}}} %

\newcommand{\ball}{\mathcal{B}} %

\newcommand{\operator}[1]{\mathcal{T}{#1}} %
\newcommand{\oparg}[2]{(\operator{#1})({#2})} %

\newcommand{\diffusion}[1]{\mathcal{T}{#1}} %
\newcommand{\diffarg}[2]{(\diffusion{#1})({#2})} %

\newcommand{\langevin}[1]{\mathcal{T}_P{#1}} %

\newcommand{\langcomp}[2]{\mathcal{T}_P^{#1}{#2}} %
\newcommand{\langarg}[2]{(\langevin{#1})({#2})} %

\newcommand{\ipm}{d_\hset} %
\newcommand{\stein}[3]{\mathcal{S}({#1},{#2},{#3})} %
\newcommand{\opstein}[2]{\stein{#1}{\operator{}}{#2}} %
\newcommand{\langstein}[2]{\stein{#1}{\langevin{}}{#2}} %
\newcommand{\wass}{d_{\wassset}} %
\newcommand{\twowass}{d_{\twowassset}} %
\newcommand{\bl}{d_{\blset}} %
\newcommand{\twobl}{d_{\twoblset}} %

\newcommand{\qvar}[0]{x} %
\newcommand{\pvar}[0]{z}%
\newcommand{\QVAR}[0]{\MakeUppercase{\qvar}} %
\newcommand{\PVAR}[0]{\MakeUppercase{\pvar}} %

\newcommand{\Vol}[1]{\textnormal{Vol}(#1)} %

\begin{document}

\twocolumn[
\icmltitle{Measuring Sample Quality with Kernels}
\icmlsetsymbol{equal}{*}

\begin{icmlauthorlist}
\icmlauthor{Jackson Gorham}{stan}
\icmlauthor{Lester Mackey}{msr}
\end{icmlauthorlist}

\icmlaffiliation{stan}{Stanford University, Palo Alto, CA USA}
\icmlaffiliation{msr}{Microsoft Research New England, Cambridge, MA USA}

\icmlcorrespondingauthor{Jackson Gorham}{jgorham@stanford.edu}
\icmlcorrespondingauthor{Lester Mackey}{lmackey@microsoft.com}

\icmlkeywords{Kernel Stein discrepancy, sample quality, reproducing kernel
Hilbert space, Stein's method, Markov chain Monte Carlo}

\vskip 0.3in
]

\printAffiliationsAndNotice{}

\begin{abstract}
Approximate Markov chain Monte Carlo (MCMC) offers the promise of more rapid sampling at the cost of more biased inference.
Since standard MCMC diagnostics fail to detect these biases, researchers have 
developed
computable \emph{Stein discrepancy} measures that provably determine %
the convergence of a sample to its target distribution.
This approach
was recently combined with the theory of reproducing kernels
to define a closed-form \emph{kernel Stein discrepancy} (KSD) computable by summing kernel evaluations across pairs of sample points. 
We develop a theory of weak convergence for KSDs based on Stein's method, 
demonstrate that commonly used KSDs fail to detect non-convergence even for Gaussian targets, 
and show that kernels with slowly decaying tails provably
determine convergence for a large class of target distributions.  The
resulting convergence-determining KSDs are suitable for comparing biased,
exact, and deterministic sample sequences and simpler to compute and
parallelize than alternative Stein discrepancies. 
We use our tools to compare biased samplers, select sampler hyperparameters,
and improve upon existing KSD approaches to one-sample hypothesis testing and sample quality improvement.
\end{abstract}

\section{Introduction}
\label{sec:intro}

When Bayesian inference and maximum likelihood
estimation \cite{Geyer91} demand the evaluation of intractable expectations
$\Esubarg{P}{h(\PVAR)} = \int p(x) h(x) dx$ under a target distribution $P$, Markov chain Monte Carlo (MCMC) methods
\cite{BrooksGeJoMe11} are often employed to approximate these 
integrals with asymptotically correct sample averages $\Esubarg{Q_n}{h(\QVAR)} = \frac{1}{n}\sum_{i=1}^n h(x_i)$.
However, many exact MCMC methods are computationally expensive, and
recent years have seen the introduction of biased MCMC procedures
\citep[see, e.g.,][]{WellingTe11,Ahn2012,Korattikara2014} that exchange asymptotic correctness
for increased sampling speed. 

Since standard MCMC diagnostics, like mean and trace plots, 
pooled and within-chain variance measures, effective sample size, and asymptotic variance~\cite{BrooksGeJoMe11}, do not account for asymptotic bias,
\citet{GorhamMa15} 
defined a new family of sample quality measures -- the \emph{Stein discrepancies} --
that measure how well $\Esub{Q_n}$ approximates $\Esub{P}$ while avoiding explicit integration under $P$.
\citet{GorhamMa15,MackeyGo16,GorhamDuVoMa16} further showed that specific members of this family -- the \emph{graph Stein discrepancies} -- were (a) efficiently computable by solving a linear program and (b) convergence-determining for large classes of targets $P$. 
Building on the zero mean reproducing kernel theory of \citet{OatesGiCh2016},
\citet{ChwialkowskiStGr2016} and \citet{LiuLeJo16} later showed that other members of the Stein discrepancy family had a closed-form solution involving the sum of kernel evaluations over pairs of sample points.

This closed form represents a significant practical advantage, as no linear program
solvers are necessary, and the computation of the discrepancy can
be easily parallelized. However, as we will see in \secref{lower-bound-ksd},
not all \emph{kernel Stein discrepancies} are suitable for our setting. 
In particular, in dimension $d\geq 3$, the kernel Stein discrepancies previously recommended in the literature fail to detect when a sample is not converging to the target. To address this shortcoming, we develop a theory of weak convergence
for the kernel Stein discrepancies analogous to that of \citep{GorhamMa15,MackeyGo16,GorhamDuVoMa16} and design a class of kernel Stein discrepancies that provably
control weak convergence for a large class of target distributions.

After formally describing our goals for measuring sample quality in \secref{quality}, we outline our strategy, based on Stein's method, for constructing and analyzing practical quality measures at the start of \secref{kernels}. In
\secref{stein_set_operator}, we define our family of closed-form quality measures -- the kernel Stein discrepancies (KSDs) -- and establish several appealing practical properties of these measures. 
We analyze the convergence properties of KSDs in \secsref{lower-bound-ksd} and \secssref{upper-bound-ksd},
showing that previously proposed KSDs fail to detect non-convergence and proposing practical convergence-determining alternatives.
\secref{experiments} illustrates the value of convergence-determining kernel Stein
discrepancies in a variety of applications, including hyperparameter selection, 
sampler selection, one-sample hypothesis testing, and sample quality improvement. 
Finally, in \secref{conclusion}, we conclude with a discussion of related and future work.

\textbf{Notation}\quad
We will use $\mu$ to denote a generic probability measure and $\Rightarrow$ to denote the weak convergence of a sequence of
probability measures.
We will use $\norm{\cdot}_r$ for $r\in [1,\infty]$
to represent the $\ell^r$ norm on $\reals^d$ 
and occasionally refer to a generic norm $\norm{\cdot}$ with
associated dual norm $\norm{a}^* \defeq \sup_{b\in \reals^d,\norm{b}=1}{\inner{a}{b}}$ for
vectors $a\in\reals^d$. We let $e_j$ be the $j$-th standard basis vector. 
For any function $g:\reals^d\to\reals^{d'}$, we define 
$M_0(g)\defeq \sup_{x\in\reals^d} \twonorm{g(x)}$, 
$M_1(g)\defeq
\sup_{x\neq y} \twonorm{g(x) - g(y)}/\twonorm{x-y}$,
and $\grad g$ as the gradient with
components $(\grad g(x))_{jk} \defeq \grad_{x_k} g_j(x)$.
We further let $g\in C^m$ indicate that $g$ is $m$ times continuously differentiable 
and $g\in C_0^m$ indicate that $g\in C^m$ and $\grad^l g$ is vanishing at infinity for all $l \in \{0,\dots,m\}$.
We define $C^{(m,m)}$ (respectively, $C_b^{(m,m)}$ and $C_0^{(m,m)}$) to be
the set of functions $k : \reals^d\times\reals^d \to \reals$ with
$(x,y) \mapsto \grad_x^l \grad_y^l k(x,y)$ continuous (respectively,
continuous and uniformly bounded, continuous and vanishing at infinity) for
all $l \in \{0,\dots,m\}$.

\section{Quality measures for samples}
\label{sec:quality}

Consider a target distribution $P$ with continuously differentiable (Lebesgue) density
$p$ supported on all of $\reals^d$.
We assume that the \emph{score function} $b\defeq \grad\log p$ can be evaluated\footnote{No knowledge of the normalizing constant is needed.} but that, for most functions of interest, direct integration under $P$ is infeasible.
We will therefore approximate integration under $P$ using a \emph{weighted sample} $Q_n = \sum_{i=1}^n q_n(x_i)\delta_{x_i}$ with sample points $x_1, \dots, x_n \in\reals^d$ and $q_n$ a probability mass function. 
We will make no assumptions about
the origins of the sample points; they may be the output of a Markov chain
or even deterministically generated.

Each $Q_n$ offers an approximation $\Esubarg{Q_n}{h(\QVAR)}
= \sum_{i=1}^n q_n(x_i) h(x_i)$
for each intractable expectation $\Esubarg{P}{h(\PVAR)}$, and
our aim is to effectively compare the quality of the approximation offered by any two samples 
targeting $P$. 
In particular, we wish to produce a quality measure that
\begin{enumerate*}[label=\emph{(\roman*)}]
\item\label{desiderata:det-conv} identifies when a sequence of samples is converging to the target,
\item\label{desiderata:det-nonconv} determines when a sequence of samples is not converging to the target, and
\item\label{desiderata:comp} is efficiently computable.
\end{enumerate*}
Since our interest is in approximating expectations, 
we will consider discrepancies quantifying the maximum expectation error over a class of test functions $\hset$:
\balign\label{eqn:ipm-definition}
d_{\hset}(Q_n, P) \defeq \sup_{h\in\hset}|\Esubarg{P}{h(\PVAR)} - \Esubarg{Q_n}{h(\QVAR)}|.
\ealign
When $\hset$ is large enough, for any sequence of probability measures
$(\mu_m)_{m\ge 1}$, $\ipm(\mu_m,P)\to 0$ only if $\mu_m\Rightarrow P$.  In
this case, we call \eqnref{ipm-definition} an \emph{integral probability
metric} (IPM) \cite{Muller97}. For example, when
$\hset=\twoblset\defeq \{h:\reals^d\to\reals\,|\, M_0(h) + M_1(h)\le 1\}$, the
IPM $\twobl$ is called the \emph{bounded Lipschitz} or \emph{Dudley metric} and exactly metrizes convergence in
distribution. Alternatively, when
$\hset=\twowassset\defeq \{h:\reals^d\to\reals\,|\, M_1(h)\le 1\}$ is the set
of $1$-Lipschitz functions, the IPM $\wass$ in \eqnref{ipm-definition} is known as
the Wasserstein metric.

An apparent practical problem with using the IPM $\ipm$ as a sample quality measure
is that $\Esubarg{P}{h(\PVAR)}$ 
may not be computable for $h\in\hset$. However, if
$\hset$ were chosen such that $\Esubarg{P}{h(\PVAR)} = 0$ for all
$h\in\hset$, then no explicit integration under $P$ would be necessary. 
To generate such a class of test functions and to show that the resulting IPM
still satisfies our desiderata, we follow the lead of \citet{GorhamMa15} and consider
Charles Stein's method for characterizing distributional convergence.

\section{Stein's method with kernels}
\label{sec:kernels}

Stein's method \cite{Stein72} provides a three-step recipe for assessing
convergence in distribution:
\benumerate
\item Identify a \emph{Stein operator} $\operator{}$ that maps functions
$g:\reals^d\to\reals^d$ from a domain $\gset$ to real-valued functions
$\operator{g}$ such that
\baligns
\Esubarg{P}{\oparg{g}{\PVAR}} = 0\text{ for all }g\in\gset.
\ealigns
For any such Stein operator and \emph{Stein set} $\gset$, \citet{GorhamMa15} defined the \emph{Stein
discrepancy} as
\balign\label{eqn:stein-discrepancy-definition}
\opstein{\mu}{\gset}
  \defeq \sup_{g\in\gset} |\Esubarg{\mu}{\oparg{g}{\QVAR}}|
  = d_{\operator{\gset}}(\mu, P)
\ealign
which, crucially, avoids explicit integration under $P$.
\item Lower bound the Stein discrepancy by an IPM $\ipm$ known to dominate weak convergence. 
This can be done once for a broad class of target distributions to ensure that %
$\mu_m \Rightarrow P$ whenever $\opstein{\mu_m}{\gset}\to 0$ for a sequence of probability measures $(\mu_m)_{m\ge 1}$  (Desideratum \ref{desiderata:det-nonconv}).
\item Provide an upper bound on the Stein discrepancy ensuring that $\opstein{\mu_m}{\gset}\to 0$
under suitable convergence of $\mu_m$ to $P$ (Desideratum \ref{desiderata:det-conv}).
\eenumerate
While Stein's method is principally used as a mathematical tool to prove convergence in distribution,
we seek, in the spirit of \cite{GorhamMa15,GorhamDuVoMa16}, to harness the Stein discrepancy as a practical tool
for measuring sample quality.
The subsections to follow develop a specific, practical instantiation of the abstract Stein's method recipe based on reproducing kernel Hilbert spaces.
An empirical analysis of the Stein discrepancies recommended by our theory follows in \secref{experiments}.

\subsection{Selecting a Stein operator and a Stein set}\label{sec:stein_set_operator}
A standard, widely applicable univariate Stein operator is the \emph{density method operator} 
\citep[see][]{SteinDiHoRe04,ChatterjeeSh11,ChenGoSh11,LeyReSw2017},
\baligns\textstyle
\oparg{g}{x} 
	\defeq \frac{1}{p(x)}\deriv{}{x}(p(x) g(x))
	= g(x)b(x) + g'(x).
\ealigns
Inspired by the generator method of \citet{Barbour88,Barbour90} and \citet{Gotze91}, 
\citet{GorhamMa15} generalized this operator to multiple dimensions.
The resulting \emph{Langevin Stein operator}
\baligns\textstyle
\langarg{g}{x} 
	\defeq \frac{1}{p(x)}\inner{\grad}{p(x) g(x)}
	= \inner{g(x)}{b(x)} + \inner{\grad}{g(x)}
\ealigns
for functions $g:\reals^d\to\reals^d$ was independently developed, without connection to Stein's method, by \citet{OatesGiCh2016} for the design of Monte Carlo control functionals.
Notably, the Langevin Stein operator depends on $P$ only through its score function
$b = \grad \log p$ and hence is computable even when the normalizing constant of $p$
is not.
While our work is compatible with other practical Stein operators, like the family
of diffusion Stein operators defined in \cite{GorhamDuVoMa16},
we will focus on the Langevin operator for the sake of brevity.

Hereafter, we will let $k:\reals^d\times\reals^d\to\reals$
be the reproducing kernel
of a reproducing kernel Hilbert space (RKHS) $\kset_k$ of functions from 
$\reals^d\to\reals$.
That is,
$\kset_k$ is a Hilbert space of functions such that, for all $x\in\reals^d$, 
$k(x,\cdot) \in \kset_k$ and $f(x) = \inner{f}{k(x,\cdot)}_{\kset_k}$ whenever
$f\in\kset_k$. We let $\knorm{\cdot}$ be the norm induced from the inner
product on $\kset_k$.

With this definition, we define our \emph{kernel Stein set}
$\ksteinsetnorm{k}{\norm{\cdot}}$ as the set of vector-valued functions $g =
(g_1, \dots, g_d)$ such that each component function $g_j$ belongs to
$\kset_k$ and the vector of their norms $\knorm{g_j}$
belongs to the $\norm{\cdot}^*$ unit ball:\footnote{Our analyses and algorithms support each $g_j$ belonging to a different RKHS $\kset_{k_j}$, but we will not need that flexibility here.}
\baligns
\ksteinsetnorm{k}{\norm{\cdot}} \defeq \{ g = (g_1,\dots, g_d) \mid \norm{v}^*\le 1 \text{ for } v_j \defeq \knorm{g_j} \}.
\ealigns
The following result, proved in \secref{stein-class-proof}, establishes that
this is an acceptable domain for $\langevin{}$.
\begin{proposition}[Zero mean test functions]\label{prop:stein-class}
If $k\in C_b^{(1,1)}$ and $\Esubarg{P}{\twonorm{\grad\log p(Z)}} < \infty$, then 
$\Esubarg{P}{\langarg{g}{Z}} = 0$ for all
$g\in\ksteinsetnorm{k}{\norm{\cdot}}$.
\end{proposition}
The Langevin Stein operator and kernel Stein set together define
our quality measure of interest, the \emph{kernel Stein discrepancy} (KSD)
$\langstein{\mu}{\ksteinsetnorm{k}{\norm{\cdot}}}$.  
When $\norm{\cdot} = \twonorm{\cdot}$, 
this definition recovers the KSD proposed by \citet{ChwialkowskiStGr2016} and \citet{LiuLeJo16}.
Our next result shows that, for any $\norm{\cdot}$, the KSD admits a
closed-form solution.
\begin{proposition}[KSD closed form]\label{prop:kernel-stein-discrepancy-formula}
Suppose $k\in C^{(1,1)}$, and, for each $j \in \{1,\dots d\}$, define the \emph{Stein kernel}
\balign
\label{eqn:k0-definition}
k_0^j(x,y) 
	&\defeq \textfrac{1}{ p(x)p(y)}\grad_{x_j} \grad_{y_j} ( p(x) k(x,y) p(y) ) \\  \notag
	&= b_j(x)b_j(y)k(x,y) + b_j(x)\grad_{y_j}k(x,y) \\ \notag
  &\quad + b_j(y)\grad_{x_j}k(x,y) + \grad_{x_j}\grad_{y_j}k(x,y).
\ealign
If $\sum_{j=1}^d\Esubarg{\mu}{{k_0^j(X,X)}^{1/2}} < \infty$, 
then $\langstein{\mu}{\ksteinsetnorm{k}{\norm{\cdot}}} =
\norm{w}$
where
$
w_j\defeq \sqrt{\Esub{\mu\times \mu}[k_0^j(X, \tilde{X})}]
$
with $X,\tilde X \iid \mu$.
\end{proposition}
The proof is found in \secref{kernel-stein-discrepancy-formula-proof}.
Notably, when $\mu$ is the discrete measure $Q_n = \sum_{i=1}^n q_n(x_i)\delta_{x_i}$, the KSD
reduces to evaluating each $k_0^j$ at pairs of support points as
$
w_j = \sqrt{\sum_{i, i'=1}^n q_n(x_i) k_0^j(x_i, x_{i'})q_n(x_{i'})},
$
a computation which is easily parallelized over sample pairs and coordinates $j$.

Our Stein set choice was motivated by the work of \citet{OatesGiCh2016} who used the sum of Stein kernels $k_0=\sum_{j=1}^d k_0^j$
to develop nonparametric control variates.
Each term $w_j$ in \propref{kernel-stein-discrepancy-formula} can also be viewed as an
instance of the maximum mean discrepancy (MMD) \cite{Gretton2012MMD} between $\mu$ and $P$ measured with respect to the Stein kernel $k_0^j$.  
In standard uses of MMD, an arbitrary kernel function is selected, and one must be able to compute expectations of the kernel function under $P$.
Here, this requirement is satisfied automatically, since our induced kernels are chosen to have mean zero under $P$.

For clarity we will focus on the specific kernel Stein set choice
$\ksteinset{k} \defeq \ksteinsetnorm{k}{\twonorm{\cdot}}$ for the remainder of the paper, 
but our results extend directly to KSDs based on any $\norm{\cdot}$, since all KSDs
are equivalent in a strong sense:
\begin{proposition}[Kernel Stein set equivalence]\label{prop:stein-set-equivalence}
Under the assumptions of \propref{kernel-stein-discrepancy-formula},
there are constants $c_d, c_d' > 0$ 
depending only on $d$ and $\norm{\cdot}$ such that
 $c_d \langstein{\mu}{\ksteinsetnorm{k}{\norm{\cdot}}}
\le \langstein{\mu}{\ksteinsetnorm{k}{\twonorm{\cdot}}}
\le c_d' \langstein{\mu}{\ksteinsetnorm{k}{\norm{\cdot}}}$.
\end{proposition}
The short proof is found in \secref{stein-set-equivalence-proof}.

\subsection{Lower bounding the kernel Stein discrepancy}\label{sec:lower-bound-ksd}
We next aim to establish conditions under which the KSD $\langstein{\mu_m}{\ksteinset{k}}\to 0$
only if $\mu_m\Rightarrow P$ (Desideratum \ref{desiderata:det-nonconv}).
Recently, \citet{GorhamDuVoMa16} showed that the Langevin graph Stein
discrepancy dominates convergence in distribution whenever $P$ belongs to the class $\pset$
of \emph{distantly dissipative} distributions with Lipschitz score function $b$:
\begin{definition}[Distant dissipativity~\cite{Eberle2015,GorhamDuVoMa16}]
A distribution $P$ is  \emph{distantly dissipative} if $\kappa_0 \defeq \lim\inf_{r\to\infty} \kappa(r) > 0$ for
\balign\label{eqn:distantly-dissipative}
\kappa(r) = \inf \{ -2 \textfrac{\inner{b(x)-b(y)}{x-y}}{\twonorm{x-y}^2} : \twonorm{x-y} = r \} .
\ealign
\end{definition}
Examples of distributions in $\pset$ include finite Gaussian mixtures with common covariance and all distributions strongly log-concave outside of a compact set, including Bayesian linear, logistic, and Huber regression posteriors with Gaussian priors~\citep[see][Section 4]{GorhamDuVoMa16}.
Moreover, when $d=1$, membership in $\pset$ is sufficient to provide a lower bound on the KSD
for most common kernels including the Gaussian, \Matern, and inverse multiquadric kernels.
\begin{theorem}[Univariate KSD detects non-convergence]\label{thm:univariate-weak-convergence}
Suppose that $P\in\pset$ and $k(x,y) = \Phi(x-y)$ for $\Phi \in C^2$ with a non-vanishing generalized Fourier transform.
If $d = 1$, then $\langstein{\mu_m}{\ksteinset{k}} \to 0$ only if $\mu_m \Rightarrow P$.
\end{theorem}

The proof in \secref{univariate-weak-convergence-proof} provides a lower bound on the KSD in terms of an IPM known to dominate weak convergence.
However, our next theorem shows that in higher dimensions
$\langstein{Q_n}{\ksteinset{k}}$ can converge to $0$ without the sequence
$(Q_n)_{n\geq1}$ converging to any probability measure.
This deficiency occurs even when the target is Gaussian.
\begin{theorem}[KSD fails with light kernel tails]\label{thm:counterexample-theorem}
Suppose $k\in C_b^{(1,1)}$ and define the kernel decay rate
\baligns
\gamma(r) \defeq \sup\{\max(&|k(x,y)|, \twonorm{\grad_x k(x,y)}, \\
& |\inner{\grad_x}{\grad_y k(x,y)}|) : \twonorm{x-y} \geq r\}.
\ealigns
If $d\geq 3$, $P = \Gsn(0,I_d)$, and $\gamma(r) = o(r^{-\alpha})$ for $\alpha \defeq (\half - \frac{1}{d})^{-1}$,
then $\langstein{Q_n}{\ksteinset{k}} \to 0$ does not imply $Q_n \Rightarrow P$.
\end{theorem}
\thmref{counterexample-theorem} implies that KSDs based on the commonly used
Gaussian kernel, \Matern kernel, and compactly supported kernels of
\citet[Theorem 9.13]{Wendland2004} all fail to detect non-convergence when $d \ge 3$. 
In addition, KSDs based on the inverse multiquadric kernel ($k(x,y) = (c^2 + \twonorm{x-y}^2)^{\beta}$) for $\beta < -1$ fail to detect non-convergence 
for any $d > 2\beta/(\beta+1)$.
The proof in \secref{counterexample-theorem-proof} shows that the 
violating sample sequences $(Q_n)_{n\geq 1}$ are simple to construct, 
and we provide an empirical demonstration of this failure to detect non-convergence in
 \secref{experiments}. 

The failure of the KSDs in \thmref{counterexample-theorem} can be traced to their
inability to enforce \emph{uniform tightness}.
A sequence of probability measures $(\mu_m)_{m\ge
  1}$ is uniformly tight if for every $\eps > 0$, there is a finite number
$R(\eps)$ such that $\lim\sup_{m} \mu_m(\twonorm{\QVAR} > R(\eps)) \le
\eps$.  Uniform tightness implies that no mass in the
sequence of probability measures escapes to infinity.
When the kernel $k$ decays more rapidly than the score function grows, 
the KSD ignores excess mass in the tails and hence can be driven to zero 
by a non-tight sequence of increasingly diffuse probability measures.
The following theorem demonstrates uniform tightness is the missing piece to
ensure weak convergence.

\begin{theorem}[KSD detects tight non-convergence]\label{thm:tightness-density-implies-weak-convergence}
Suppose that $P\in\pset$ and $k(x,y) = \Phi(x-y)$ for $\Phi \in C^2$ with a non-vanishing generalized Fourier transform.
If $(\mu_m)_{m\geq 1}$ is uniformly tight, 
then $\langstein{\mu_m}{\ksteinset{k}} \to 0$ only if $\mu_m \Rightarrow P$.
\end{theorem}

Our proof in \secref{tightness-density-implies-weak-convergence-proof}
explicitly lower bounds the KSD
$\langstein{\mu}{\ksteinset{k}}$ in terms of the bounded Lipschitz metric $\bl(\mu,P)$, which exactly metrizes weak convergence.

Ideally, when a sequence of probability measures is not uniformly tight, the
KSD would reflect this divergence in its reported value. To achieve this, we consider
the inverse multiquadric (IMQ) kernel
$%
k(x,y) = (c^2 + \twonorm{x-y}^2)^{\beta}\, %
$ %
for some $\beta < 0$ and $c > 0$.
While KSDs based on IMQ kernels fail to determine convergence when $\beta < -1$ (by \thmref{counterexample-theorem}), our next theorem
shows that they automatically enforce tightness and detect non-convergence whenever
$\beta\in (-1,0)$.

\begin{theorem}[IMQ KSD detects non-convergence]\label{thm:imq-implies-weak-convergence}
Suppose $P\in\pset$ and $k(x,y) = (c^2 + \twonorm{x-y}^2)^{\beta}$ for $c>0$ and
$\beta\in (-1,0)$. If $\langstein{\mu_m}{\ksteinset{k}} \to 0$, then $\mu_m
\Rightarrow P$.
\end{theorem}

The proof in \secref{imq-implies-weak-convergence-proof} provides a
lower bound on the KSD in terms of the bounded Lipschitz metric $\bl(\mu, P)$. 
The success of the IMQ kernel over other common characteristic kernels can be attributed to its slow decay
rate. When $P\in\pset$ and the IMQ exponent $\beta > -1$, the function class $\langevin{\ksteinset{k}}$
contains unbounded (coercive) functions. These functions ensure that the IMQ KSD $\langstein{\mu_m}{\ksteinset{k}}$
goes to $0$ only if $(\mu_m)_{m\geq 1}$ is uniformly tight.

\subsection{Upper bounding the kernel Stein discrepancy}\label{sec:upper-bound-ksd}
The usual goal in upper bounding the Stein discrepancy is to provide a rate
of convergence to $P$ for particular approximating sequences
$(\mu_m)_{m=1}^\infty$. Because we aim to directly compute the KSD for arbitrary 
samples $Q_n$, our chief purpose in this section is to ensure that
the KSD $\langstein{\mu_m}{\ksteinset{k}}$ will converge to zero when
$\mu_m$ is converging to $P$ (Desideratum \ref{desiderata:det-conv}).

\begin{proposition}[KSD detects convergence]\label{prop:ksd-upper-bound}
If $k \in C_b^{(2,2)}$ and $\grad \log p$ is Lipschitz
with $\Esub{P}[\twonorm{\grad \log p(Z)}^2] < \infty$,
then
$\langstein{\mu_m}{\ksteinset{k}}\to 0$ whenever
the Wasserstein distance $\twowass(\mu_m,P)\to 0$.
\end{proposition}
\propref{ksd-upper-bound} applies to common kernels like the Gaussian, \Matern, and IMQ kernels, and its proof in \secref{ksd-upper-bound-proof} provides an explicit upper bound on the KSD in terms of the Wasserstein distance $\twowass$.
When $Q_n = \frac{1}{n}\sum_{i=1}^n \delta_{x_i}$ for $x_i \iid \mu$, 
\citep[Thm. 4.1]{LiuLeJo16} further implies that $\langstein{Q_n}{\ksteinset{k}} \Rightarrow \langstein{\mu}{\ksteinset{k}}$
at an $O(n^{-1/2})$ rate under continuity and integrability assumptions on $\mu$.

\section{Experiments}
\label{sec:experiments}

We next conduct an empirical evaluation of the KSD quality measures
recommended by our theory, recording all timings on an Intel Xeon CPU E5-2650 v2 @ 2.60GHz.  
Throughout, we will refer to the KSD with IMQ
base kernel $k(x,y) = (c^2 + \twonorm{x-y}^2)^{\beta}$, exponent
$\beta=-\half$, and $c=1$ as the IMQ KSD.  
Code reproducing all experiments can be found on the Julia \cite{Bezanson2014julia} 
package site \url{https://jgorham.github.io/SteinDiscrepancy.jl/}.

\subsection{Comparing discrepancies}
\label{sec:comparing-discrepancies}
\begin{figure*}
  \centering
  \includegraphics[width=\textwidth]{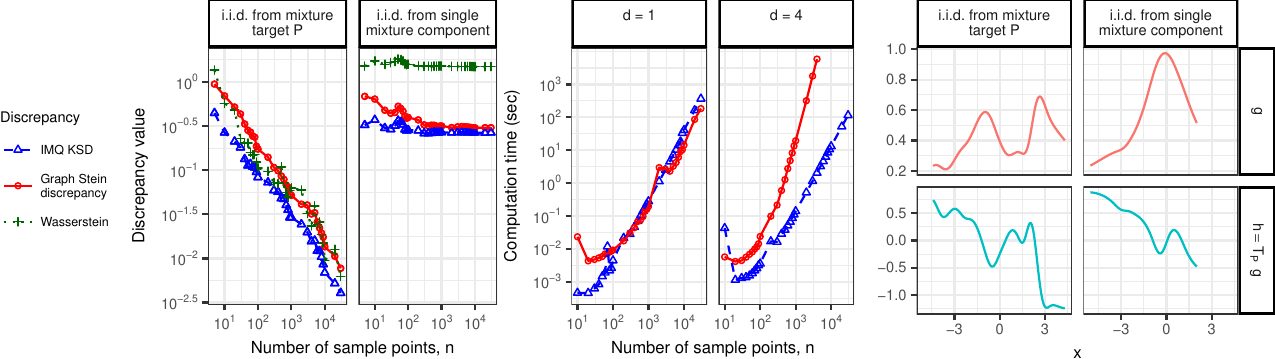}
  \caption{
  \textbf{Left:} For $d=1$, comparison of discrepancy measures for samples drawn i.i.d.\ from either the bimodal Gaussian mixture target $P$ or a single mixture component (see \secref{comparing-discrepancies}).
    \textbf{Middle:} On-target discrepancy computation time using $d$ cores in $d$ dimensions.
    \textbf{Right:} For $n=10^3$ and $d=1$,
    the Stein functions $g$ and discriminating test functions $h=\langevin{g}$ which maximize the KSD.
  }
  \label{fig:kernel_vs_graph}
\end{figure*}

Our first, simple experiment is designed to illustrate several properties of
the IMQ KSD and to compare its behavior with that of two preexisting
discrepancy measures, the Wasserstein distance $\twowass$, which can be
computed for simple univariate targets~\citep{Vallender74}, and the spanner
graph Stein discrepancy of \citet{GorhamMa15}.  We adopt a bimodal Gaussian
mixture with
$p(x)\propto e^{-\half \twonorm{x+\Delta e_1}^2} +
e^{-\half \twonorm{x - \Delta e_1}^2}$ and $\Delta = 1.5$ as our target $P$ 
and generate a first sample point sequence i.i.d.\ from the target and a
second sequence i.i.d.\ from one component of the mixture, $\Gsn(-\Delta e_1,
I_d)$. As seen in the left panel of \figref{kernel_vs_graph} where $d=1$,
the IMQ KSD decays at an $n^{-0.51}$ rate when applied to the first $n$
points in the target sample and remains bounded away from zero when applied
to the to the single component sample.  This desirable behavior is closely
mirrored by the Wasserstein distance and the graph Stein discrepancy.

The middle panel of \figref{kernel_vs_graph} records the time consumed by
the graph and kernel Stein discrepancies applied to the i.i.d.\ sample
points from $P$.  Each method is given access to $d$ cores when working in
$d$ dimensions, and we use the released code of \citet{GorhamMa15} with the
default Gurobi 6.0.4 linear program solver for the graph Stein discrepancy.
We find that the two methods have nearly identical runtimes when $d=1$ but
that the KSD is $10$ to $1000$ times faster when $d=4$.  In addition, the
KSD is straightforwardly parallelized and does not require access to a
linear program solver, making it an appealing practical choice for a quality
measure.

Finally, the right panel displays the optimal Stein functions,
$g_j(y) = \frac{\Esubarg{Q_n}{b_j(X)k(X,y) + \grad_{x_j}k(X,y)}}{\langstein{Q_n}{\ksteinset{k}}}$, recovered by the IMQ KSD when $d=1$ and $n=10^3$.
The associated test functions $h(y) =\langarg{g}{y} = \frac{\sum_{j=1}^d \Esubarg{Q_n}{k_0^j(X,y)}}{\langstein{Q_n}{\ksteinset{k}}}$ are the mean-zero functions
under $P$ that best discriminate the target $P$ and the sample $Q_n$.
The optimal test function for the single component sample features large
positive values in the oversampled region that fail to be offset by negative
values in the undersampled region near the missing mode.

\subsection{The importance of kernel choice}
\label{sec:importance-of-kernel-choice}
\thmref{counterexample-theorem} established that kernels
with rapidly decaying tails yield KSDs that can be driven to zero by off-target sample sequences.  
Our next experiment provides an empirical demonstration of this issue for a
multivariate Gaussian target $P = \Gsn(0,I_d)$ and KSDs based on the popular
Gaussian ($k(x,y) = e^{-\twonorm{x-y}^2/2}$) and \Matern ($k(x,y) =
(1+\sqrt{3}\twonorm{x-y})e^{-\sqrt{3}\twonorm{x-y}}$) radial kernels.

Following the proof \thmref{counterexample-theorem}
in \secref{counterexample-theorem-proof}, we construct an off-target
sequence $(Q_n)_{n\ge 1}$ that sends $\langstein{Q_n}{\ksteinset{k}}$ to
$0$ for these kernel choices whenever $d \geq 3$.  Specifically, for each $n$,
we let $Q_n = \frac{1}{n}\sum_{i=1}^n \delta_{x_i}$ where, for all $i$ and
$j$, $\twonorm{x_i} \leq 2n^{1/d}\log n $ and $\twonorm{x_i - x_j} \geq
2\log n$.  To select these sample points, we independently sample candidate
points uniformly from the ball $\{x : \twonorm{x} \leq 2n^{1/d}\log n\}$,
accept any points not within $2\log n$ Euclidean distance of any previously
accepted point, and terminate when $n$ points have been accepted.

\begin{figure}[t]
  \centering
  \includegraphics[width=0.5\textwidth]{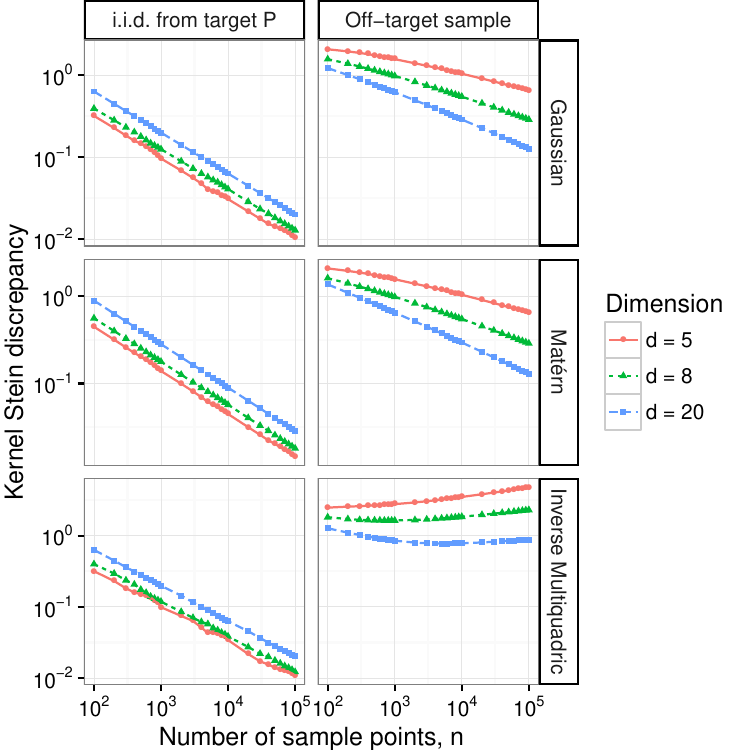}
  \caption{
  Gaussian and \Matern KSDs are driven to $0$ by an off-target sequence that does not converge to the target $P = \Gsn(0,I_d)$ (see \secref{importance-of-kernel-choice}).
  The IMQ KSD does not share this deficiency.
  }
  \label{fig:compare_coercive_kernel_discrepancies_diagnostics}
\end{figure}

For various
dimensions, \figref{compare_coercive_kernel_discrepancies_diagnostics}
displays the result of applying each KSD to the off-target sequence
$(Q_n)_{n\geq 1}$ and an ``on-target'' sequence of points sampled i.i.d.\
from $P$.  For comparison, we also display the behavior of the IMQ KSD which
provably controls tightness and dominates weak convergence for this target
by \thmref{imq-implies-weak-convergence}.  As predicted, the Gaussian
and \Matern KSDs decay to $0$ under the off-target sequence and decay more
rapidly as the dimension $d$ increases; the IMQ KSD remains bounded
away from $0$.

\subsection{Selecting sampler hyperparameters}
\label{sec:hyperparameter}
The approximate slice sampler of \citet{Dubois2014} is a biased MCMC
procedure designed to accelerate inference when the target density takes the
form
$p(x) \propto \pi(x) \prod_{l=1}^L \pi(y_l | x)$ for $\pi(\cdot)$ a prior
distribution on $\reals^d$ and $\pi(y_l | x)$ the likelihood of a datapoint
$y_l$.  A standard slice sampler must evaluate the likelihood of all $L$
datapoints to draw each new sample point $x_i$.  To reduce this cost, the
approximate slice sampler introduces a tuning parameter $\eps$ which
determines the number of datapoints that contribute to an approximation of
the slice sampling step; an appropriate setting of this parameter is
imperative for accurate inference.  When $\eps$ is too small, relatively few
sample points will be generated in a given amount of sampling time, yielding
sample expectations with high Monte Carlo variance.  When $\eps$ is too
large, the large approximation error will produce biased samples that no
longer resemble the target.

\begin{figure*}[t!]
  \centering
  \includegraphics[width=\textwidth]{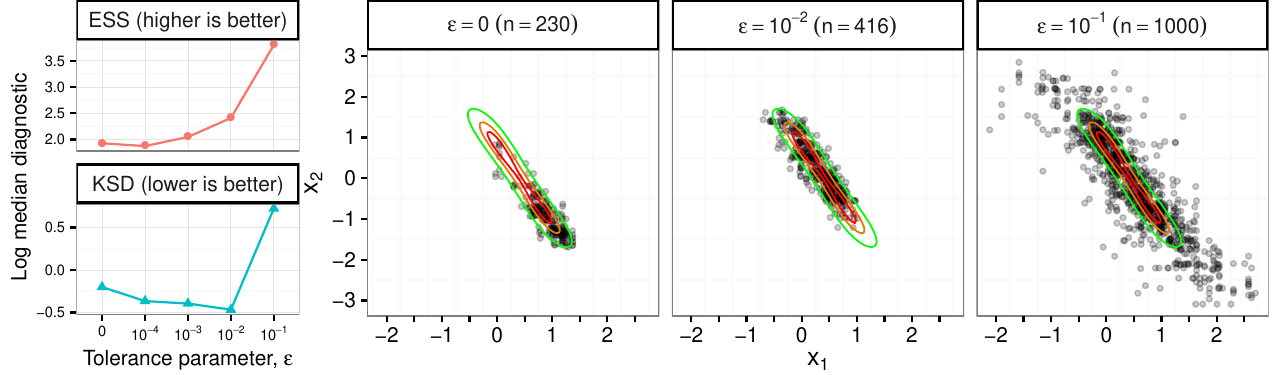}
  \caption{
  \textbf{Left:} Median hyperparameter selection criteria across $50$
  independent approximate slice sampler sample sequences
  (see \secref{hyperparameter}); IMQ KSD selects $\eps = 10^{-2}$; effective
  sample size selects $\eps = 10^{-1}$.
  \textbf{Right:} Representative approximate slice sampler samples requiring
  $148000$ likelihood evaluations with posterior equidensity contours
  overlaid; $n$ is the associated sample size.
  }
  \label{fig:compare-hyperparameters-approxslice}
\end{figure*}

To assess the suitability of the KSD for tolerance parameter selection, we
take as our target $P$ the bimodal Gaussian mixture model posterior
of \cite{WellingTe11}.  For an array of $\eps$ values, we generated $50$
independent approximate slice sampling chains with batch size $5$, each with a budget of
$148000$ likelihood evaluations, and plotted the median IMQ KSD and
effective sample size (ESS, a standard sample quality measure based on
asymptotic variance \cite{BrooksGeJoMe11})
in \figref{compare-hyperparameters-approxslice}.  ESS, which does not detect
Markov chain bias, is maximized at the largest hyperparameter evaluated
($\eps = 10^{-1}$), while the KSD is minimized at an intermediate value
($\eps = 10^{-2}$).  The right panel
of \figref{compare-hyperparameters-approxslice} shows representative samples
produced by several settings of $\eps$.  The sample produced by the
ESS-selected chain is significantly overdispersed, while the sample from
$\eps = 0$ has minimal coverage of the second mode due to its small sample
size.  The sample produced by the KSD-selected chain best resembles the
posterior target.  Using $4$ cores, the longest KSD computation with
$n=10^3$ sample points took $0.16s$.

\subsection{Selecting samplers} %
\label{sec:selecting-samplers}

\citet{Ahn2012} developed two biased MCMC samplers for accelerated posterior inference,
both called Stochastic Gradient Fisher Scoring (SGFS).  In the full version
of SGFS (termed SGFS-f), a $d\times d$ matrix must be inverted to draw each
new sample point.  Since this can be costly for large $d$, the authors
developed a second sampler (termed SGFS-d) in which only a diagonal matrix
must be inverted to draw each new sample point.  Both samplers can be viewed
as discrete-time approximations to a continuous-time Markov process that has
the target $P$ as its stationary distribution; however, because no
Metropolis-Hastings correction is employed, neither sampler has the target
as its stationary distribution.  Hence we will use the KSD -- a quality
measure that accounts for asymptotic bias -- to evaluate and choose between
these samplers.

\begin{figure*}
  \centering
  \includegraphics[width=\textwidth]{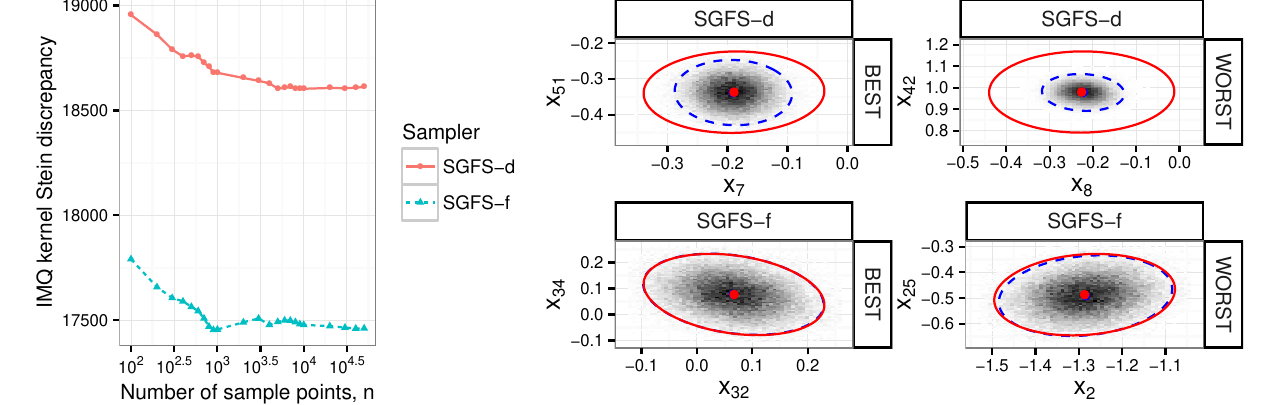}
  \caption{
  \textbf{Left:}
  Quality comparison for Bayesian logistic regression with two SGFS samplers (see \secref{selecting-samplers}).
  \textbf{Right:} Scatter plots of $n=5\times 10^4$ SGFS sample points with overlaid bivariate marginal means and 95\% confidence ellipses (dashed blue) that align best and worst with surrogate ground truth sample (solid red).
  }
  \label{fig:mnist_7_or_9_sgfs}
\end{figure*}

Specifically, we evaluate the SGFS-f and SGFS-d samples produced
in \citep[Sec. 5.1]{Ahn2012}.  The target $P$ is a Bayesian logistic
regression with a flat prior, conditioned on a dataset of $10^4$ MNIST
handwritten digit images.  From each image, the authors extracted $50$
random projections of the raw pixel values as covariates and a label
indicating whether the image was a $7$ or a $9$.  After discarding the first
half of sample points as burn-in, we obtained regression coefficient samples
with $5\times 10^4$ points and $d=51$ dimensions (including the intercept
term).
\figref{mnist_7_or_9_sgfs} displays the IMQ KSD applied to the first $n$ points in each sample.
As external validation, we follow the protocol of \citet{Ahn2012} to find the
bivariate marginal means and 95\% confidence ellipses of each sample that
align best and worst with those of a surrogate ground truth sample obtained
from a Hamiltonian Monte Carlo chain with $10^5$ iterates.  Both the KSD and
the surrogate ground truth suggest that the moderate speed-up provided by
SGFS-d ($0.0017s$ per sample vs. $0.0019s$ for SGFS-f) is outweighed by the
significant loss in inferential accuracy.  However, the KSD assessment does
not require access to an external trustworthy ground truth sample.  The
longest KSD computation took $400s$ using $16$ cores.

\subsection{Beyond sample quality comparison}

While our investigation of the KSD was motivated by the desire to develop practical, trustworthy tools for sample quality comparison, 
the kernels recommended by our theory can serve as drop-in replacements in other inferential tasks that make use of kernel Stein discrepancies.
\subsubsection{One-sample hypothesis testing}
\label{sec:hypothesis-testing}
\citet{ChwialkowskiStGr2016} recently used the KSD $\langstein{Q_n}{\ksteinset{k}}$ to develop a hypothesis test of whether a given sample from a Markov chain was drawn from a target distribution $P$ \citep[see also][]{LiuLeJo16}. 
However, the authors noted that the KSD test with their default Gaussian base kernel $k$ experienced a considerable loss of power as the dimension $d$ increased.
We recreate their experiment and show that this loss of power can be avoided by using our default IMQ kernel with $\beta = -\half$ and $c=1$.
Following \citep[Section 4]{ChwialkowskiStGr2016} we draw
$z_i\iid \Gsn(0, I_d)$ and $u_i\iid \Unif[0,1]$ to generate a sample
$(x_i)_{i=1}^{n}$ with $x_i = z_i + u_i\,e_1$ for $n=500$ and various dimensions
$d$. Using the authors' code (modified to include an IMQ kernel), we compare the power of the Gaussian KSD test, 
the IMQ KSD test, and the standard normality test of \citet{BaringHaus1988} (B\&H)
to discern whether the sample $(x_i)_{i=1}^{500}$ came from the null distribution $P = \Gsn(0,I_d)$. 
The results, averaged over $400$ simulations, are shown in \tabref{one-sample-power}. 
Notably, the IMQ KSD experiences no power degradation over this range of dimensions, thus improving on both the Gaussian KSD and the standard B\&H normality tests.

\begin{table}[h]
  \caption{Power of one sample tests for multivariate normality, averaged over $400$ simulations (see \secref{hypothesis-testing})}
  \begin{center}
    \begin{tabular}{| c | c | c | c | c | c | c |}
      \hline
         & d=2 & d=5 & d=10 & d=15 & d=20 & d=25 \\ \hline
       B\&H & 1.0 & 1.0 & 1.0 & 0.91 & 0.57 & 0.26 \\ \hline
       Gaussian & 1.0 & 1.0 & 0.88 & 0.29 & 0.12 & 0.02 \\ \hline
       IMQ & 1.0 & 1.0 & 1.0 & 1.0 & 1.0 & 1.0 \\
      \hline
    \end{tabular}
  \end{center}
  \label{tab:one-sample-power}
\end{table}

\subsubsection{Improving sample quality}
\label{sec:improving}
\begin{figure}[t]
  \centering
  \includegraphics[width=0.5\textwidth]{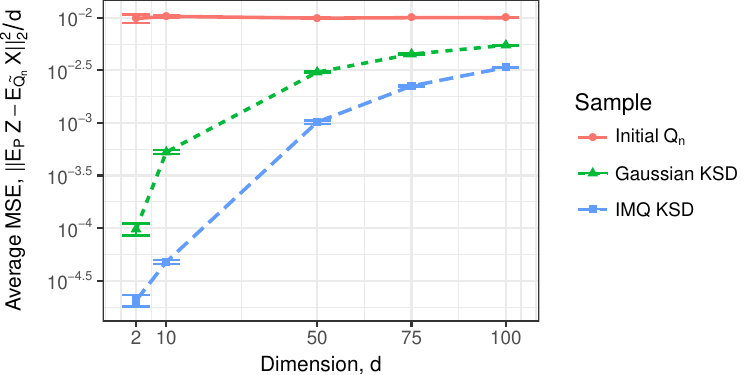}
  \caption{
  Average quality of mean estimates ($\pm 2$ standard errors) under optimized samples $\tilde{Q}_n$ for target $P = \Gsn(0,I_d)$;
  MSE averaged over $500$ independent initial samples (see \secref{improving}).
  }
  \label{fig:blackweights_n=100}
\end{figure}

\citet{LiuLe2016} recently used the KSD $\langstein{Q_n}{\ksteinset{k}}$ as a means of improving the quality of a sample.  Specifically, given an initial sample $Q_n$ supported on $x_1, \dots, x_n$, they minimize $\langstein{\tilde{Q}_n}{\ksteinset{k}}$ over all measures $\tilde{Q}_n$ supported on the same sample points to obtain a new sample that better approximates $P$ over the class of test functions $\hset = \langevin{\ksteinset{k}}$.
In all experiments, \citet{LiuLe2016} employ a Gaussian kernel $k(x,y) = e^{-\frac{1}{h} \twonorm{x-y}^2}$ 
with bandwidth $h$ selected to be the median of the squared Euclidean distance between pairs of sample points.
Using the authors' code, we recreate the experiment from \citep[Fig. 2b]{LiuLe2016} 
and introduce a KSD objective with an IMQ kernel $k(x,y) = (1+\frac{1}{h} \twonorm{x-y}^2)^{-1/2}$ with bandwidth selected in the same fashion.
The starting sample is given by $Q_n = \frac{1}{n} \sum_{i=1}^n \delta_{x_i}$ for $n=100$, various dimensions $d$, and each sample point drawn i.i.d.\ from $P = \Gsn(0,I_d)$.
For the initial sample and the optimized samples produced by each KSD, \figref{blackweights_n=100} displays the mean squared error (MSE) $\frac{1}{d}\statictwonorm{\Esubarg{P}{\PVAR} - \Esubarg{\tilde{Q}_n}{X}}^2$ averaged across $500$ independently generated initial samples.
Out of the box, the IMQ kernel produces better mean estimates than the standard Gaussian.

\section{Related and future work}
\label{sec:conclusion}

The score statistic of~\citet{FanBrGe06} and
the Gibbs sampler convergence criteria of~\citet{ZellnerMi95} detect certain
forms of non-convergence but fail to detect others due to the finite number
of test functions tested. For example, when $P = \Gsn(0,1)$, the score
statistic~\citep{FanBrGe06} only monitors sample means and variances.

For an approximation $\mu$ with continuously differentiable density $r$, \citet[Thm. 2.2]{ChwialkowskiStGr2016} and \citet[Prop. 3.3]{LiuLeJo16} established 
that if $k$ is $C_0$-universal~\citep[Defn. 4.1]{Carmeli2010} or integrally strictly positive definite~\citep[ISPD,][Sec. 6]{Stewart1976} and 
$\Esub{\mu}[k_0(X,X)+ \statictwonorm{\grad\log \frac{p(X)}{r(X)}}^2] < \infty$ for $k_0 \defeq \sum_{j=1}^d k_0^j$,
then $\langstein{\mu}{\ksteinset{k}} = 0$ only if $\mu = P$. 
However, this property is insufficient to conclude that probability measures with small
KSD are close to $P$ in any traditional sense.
Indeed, Gaussian and \Matern kernels are $C_0$ universal and ISPD,
but, by \thmref{counterexample-theorem}, their KSDs can be driven to zero by sequences not converging to $P$.
On compact domains, where tightness is no longer an issue, the combined results of \citep[Lem.~4]{Oates2016}, \citep[Lem.~1]{Fukumizu2007kernel}, and \citep[Thm.~55]{Simon2016kernel} give conditions for a KSD to dominate weak convergence.

While assessing sample quality was our chief objective, our results 
may hold benefits for other applications that make use of Stein discrepancies or Stein operators.
In particular, our kernel recommendations could be incorporated into the Monte Carlo control functionals framework 
of \citet{OatesGiCh2016, Oates2015}, the variational inference approaches of \citet{LiuWa2016,Liu2016two,Ranganath2016},
and the Stein generative adversarial network approach of \citet{WangLi2016}.

In the future, we aim to leverage stochastic, low-rank, and sparse
approximations of the kernel matrix and score function to produce
KSDs that scale better with the number of sample and data points
while still guaranteeing control over weak convergence.
A reader may also wonder for which distributions outside of
$\pset$ the KSD dominates weak convergence.
The following theorem, proved in \secref{bounded-score-function-proof},
shows that no KSD with a $C_0$ kernel dominates weak
convergence when the target has a bounded score function.
\begin{theorem}[KSD fails for bounded scores]\label{thm:bounded-score-function}
If $\grad \log p$ is bounded and $k \in C^{(1,1)}_0$,
then $\langstein{Q_n}{\ksteinset{k}}\to 0$ does not imply $Q_n\Rightarrow P$.
\end{theorem}
However, \citet{GorhamDuVoMa16} developed convergence-determining graph Stein discrepancies for  heavy-tailed targets by replacing the Langevin Stein operator $\langevin{}$ with \emph{diffusion Stein operators} of the form $\diffarg{g}{x} = \textfrac{1}{p(x)}\inner{\grad}{p(x)(a(x) + c(x))g(x)}$.
An analogous construction should yield convergence-determining \emph{diffusion KSDs} for $P$ outside of $\pset$.
Our results also extend to targets $P$ supported on a convex subset $\xset$
of $\reals^d$ by choosing $k$ to satisfy $p(x)k(x,\cdot)\equiv 0$ for all
$x$ on the boundary of $\xset$.

\ifdefined\arxivmode
\section*{Acknowledgments}
We thank Kacper Chwialkowski, Heiko Strathmann, and Arthur Gretton for sharing their hypothesis
testing code, Qiang Liu for sharing his black-box importance sampling
code, and Sebastian Vollmer and Andrew Duncan for many helpful conversations regarding this work.
This material is based upon work supported by the National Science Foundation DMS RTG Grant No. 1501767,
the National Science Foundation Graduate Research Fellowship under Grant No. DGE-114747, and the
Frederick E. Terman Fellowship.
\fi

\bibliography{stein}

\begin{thebibliography}{47}
\providecommand{\natexlab}[1]{#1}
\providecommand{\url}[1]{\texttt{#1}}
\expandafter\ifx\csname urlstyle\endcsname\relax
  \providecommand{\doi}[1]{doi: #1}\else
  \providecommand{\doi}{doi: \begingroup \urlstyle{rm}\Url}\fi

\bibitem[Ahn et~al.(2012)Ahn, Korattikara, and Welling]{Ahn2012}
Ahn, S., Korattikara, A., and Welling, M.
\newblock Bayesian posterior sampling via stochastic gradient {F}isher scoring.
\newblock In \emph{Proc. 29th ICML}, ICML'12, 2012.

\bibitem[Bachman \& Narici(1966)Bachman and Narici]{Bachman1966functional}
Bachman, G. and Narici, L.
\newblock \emph{Functional Analysis}.
\newblock Academic Press textbooks in mathematics. Dover Publications, 1966.
\newblock ISBN 9780486402512.

\bibitem[Baker(1999)]{Baker1999}
Baker, J.
\newblock Integration of radial functions.
\newblock \emph{Mathematics Magazine}, 72\penalty0 (5):\penalty0 392--395,
  1999.

\bibitem[Barbour(1988)]{Barbour88}
Barbour, A.~D.
\newblock Stein's method and {P}oisson process convergence.
\newblock \emph{J. Appl. Probab.}, \penalty0 (Special Vol. 25A):\penalty0
  175--184, 1988.
\newblock ISSN 0021-9002.
\newblock A celebration of applied probability.

\bibitem[Barbour(1990)]{Barbour90}
Barbour, A.~D.
\newblock Stein's method for diffusion approximations.
\newblock \emph{Probab. Theory Related Fields}, 84\penalty0 (3):\penalty0
  297--322, 1990.
\newblock ISSN 0178-8051.
\newblock \doi{10.1007/BF01197887}.

\bibitem[Baringhaus \& Henze(1988)Baringhaus and Henze]{BaringHaus1988}
Baringhaus, L. and Henze, N.
\newblock A consistent test for multivariate normality based on the empirical
  characteristic function.
\newblock \emph{Metrika}, 35\penalty0 (1):\penalty0 339--348, 1988.

\bibitem[Bezanson et~al.(2014)Bezanson, Edelman, Karpinski, and
  Shah]{Bezanson2014julia}
Bezanson, J., Edelman, A., Karpinski, S., and Shah, V.B.
\newblock Julia: A fresh approach to numerical computing.
\newblock \emph{arXiv preprint arXiv:1411.1607}, 2014.

\bibitem[Brooks et~al.(2011)Brooks, Gelman, Jones, and Meng]{BrooksGeJoMe11}
Brooks, S., Gelman, A., Jones, G., and Meng, X.-L.
\newblock \emph{Handbook of {M}arkov chain {M}onte {C}arlo}.
\newblock CRC press, 2011.

\bibitem[Carmeli et~al.(2010)Carmeli, De~Vito, Toigo, and
  Umanit{\'a}]{Carmeli2010}
Carmeli, C., De~Vito, E., Toigo, A., and Umanit{\'a}, V.
\newblock Vector valued reproducing kernel hilbert spaces and universality.
\newblock \emph{Analysis and Applications}, 8\penalty0 (01):\penalty0 19--61,
  2010.

\bibitem[Chatterjee \& Shao(2011)Chatterjee and Shao]{ChatterjeeSh11}
Chatterjee, S. and Shao, Q.
\newblock Nonnormal approximation by {S}tein's method of exchangeable pairs
  with application to the {C}urie-{W}eiss model.
\newblock \emph{Ann. Appl. Probab.}, 21\penalty0 (2):\penalty0 464--483, 2011.
\newblock ISSN 1050-5164.
\newblock \doi{10.1214/10-AAP712}.

\bibitem[Chen et~al.(2011)Chen, Goldstein, and Shao]{ChenGoSh11}
Chen, L., Goldstein, L., and Shao, Q.
\newblock \emph{Normal approximation by {S}tein's method}.
\newblock Probability and its Applications. Springer, Heidelberg, 2011.
\newblock ISBN 978-3-642-15006-7.
\newblock \doi{10.1007/978-3-642-15007-4}.

\bibitem[Chwialkowski et~al.(2016)Chwialkowski, Strathmann, and
  Gretton]{ChwialkowskiStGr2016}
Chwialkowski, K., Strathmann, H., and Gretton, A.
\newblock A kernel test of goodness of fit.
\newblock In \emph{Proc. 33rd ICML}, ICML, 2016.

\bibitem[DuBois et~al.(2014)DuBois, Korattikara, Welling, and
  Smyth]{Dubois2014}
DuBois, C., Korattikara, A., Welling, M., and Smyth, P.
\newblock Approximate slice sampling for {B}ayesian posterior inference.
\newblock In \emph{Proc. 17th AISTATS}, pp.\  185--193, 2014.

\bibitem[Eberle(2015)]{Eberle2015}
Eberle, A.
\newblock Reflection couplings and contraction rates for diffusions.
\newblock \emph{Probab. Theory Related Fields}, pp.\  1--36, 2015.
\newblock \doi{10.1007/s00440-015-0673-1}.

\bibitem[Fan et~al.(2006)Fan, Brooks, and Gelman]{FanBrGe06}
Fan, Y., Brooks, S.~P., and Gelman, A.
\newblock Output assessment for {M}onte {C}arlo simulations via the score
  statistic.
\newblock \emph{J. Comp. Graph. Stat.}, 15\penalty0 (1), 2006.

\bibitem[Fukumizu et~al.(2007)Fukumizu, Gretton, Sun, and
  Sch{\"o}lkopf]{Fukumizu2007kernel}
Fukumizu, K., Gretton, A., Sun, X., and Sch{\"o}lkopf, B.
\newblock Kernel measures of conditional dependence.
\newblock In \emph{NIPS}, volume~20, pp.\  489--496, 2007.

\bibitem[Geyer(1991)]{Geyer91}
Geyer, C.~J.
\newblock Markov chain {M}onte {C}arlo maximum likelihood.
\newblock \emph{Computer Science and Statistics: Proc.{ }23rd Symp.{
  }Interface}, pp.\  156--163, 1991.

\bibitem[Gorham \& Mackey(2015)Gorham and Mackey]{GorhamMa15}
Gorham, J. and Mackey, L.
\newblock Measuring sample quality with {S}tein's method.
\newblock In Cortes, C., Lawrence, N.~D., Lee, D.~D., Sugiyama, M., and
  Garnett, R. (eds.), \emph{Adv. NIPS 28}, pp.\  226--234. Curran Associates,
  Inc., 2015.

\bibitem[Gorham et~al.(2016)Gorham, Duncan, Vollmer, and
  Mackey]{GorhamDuVoMa16}
Gorham, J., Duncan, A., Vollmer, S., and Mackey, L.
\newblock Measuring sample quality with diffusions.
\newblock \emph{arXiv:1611.06972}, Nov. 2016.

\bibitem[G{\"o}tze(1991)]{Gotze91}
G{\"o}tze, F.
\newblock On the rate of convergence in the multivariate {CLT}.
\newblock \emph{Ann. Probab.}, 19\penalty0 (2):\penalty0 724--739, 1991.

\bibitem[Gretton et~al.(2012)Gretton, Borgwardt, Rasch, Sch{\"o}lkopf, and
  Smola]{Gretton2012MMD}
Gretton, A., Borgwardt, K., Rasch, M., Sch{\"o}lkopf, B., and Smola, A.
\newblock A kernel two-sample test.
\newblock \emph{J. Mach. Learn. Res.}, 13\penalty0 (1):\penalty0 723--773,
  2012.

\bibitem[Herb \& Sally~Jr.(2011)Herb and Sally~Jr.]{herb2011plancherel}
Herb, R. and Sally~Jr., P.J.
\newblock The {P}lancherel formula, the {P}lancherel theorem, and the {F}ourier
  transform of orbital integrals.
\newblock In \emph{Representation Theory and Mathematical Physics: Conference
  in Honor of Gregg Zuckerman's 60th Birthday, October 24--27, 2009, Yale
  University}, volume 557, pp.\ ~1. American Mathematical Soc., 2011.

\bibitem[Korattikara et~al.(2014)Korattikara, Chen, and
  Welling]{Korattikara2014}
Korattikara, A., Chen, Y., and Welling, M.
\newblock Austerity in {MCMC} land: Cutting the {M}etropolis-{H}astings budget.
\newblock In \emph{Proc. of 31st ICML}, ICML'14, 2014.

\bibitem[Ley et~al.(2017)Ley, Reinert, and Swan]{LeyReSw2017}
Ley, C., Reinert, G., and Swan, Y.
\newblock Stein's method for comparison of univariate distributions.
\newblock \emph{Probab. Surveys}, 14:\penalty0 1--52, 2017.
\newblock \doi{10.1214/16-PS278}.

\bibitem[Liu \& Feng(2016)Liu and Feng]{Liu2016two}
Liu, Q. and Feng, Y.
\newblock Two methods for wild variational inference.
\newblock \emph{arXiv preprint arXiv:1612.00081}, 2016.

\bibitem[{Liu} \& {Lee}(2016){Liu} and {Lee}]{LiuLe2016}
{Liu}, Q. and {Lee}, J.
\newblock Black-box importance sampling.
\newblock \emph{arXiv:1610.05247}, October 2016.
\newblock To appear in AISTATS 2017.

\bibitem[{Liu} \& {Wang}(2016){Liu} and {Wang}]{LiuWa2016}
{Liu}, Q. and {Wang}, D.
\newblock {Stein Variational Gradient Descent: A General Purpose Bayesian
  Inference Algorithm}.
\newblock \emph{arXiv:1608.04471}, August 2016.

\bibitem[Liu et~al.(2016)Liu, Lee, and Jordan]{LiuLeJo16}
Liu, Q., Lee, J., and Jordan, M.
\newblock A kernelized {S}tein discrepancy for goodness-of-fit tests.
\newblock In \emph{Proc. of 33rd ICML}, volume~48 of \emph{ICML}, pp.\
  276--284, 2016.

\bibitem[Mackey \& Gorham(2016)Mackey and Gorham]{MackeyGo16}
Mackey, L. and Gorham, J.
\newblock Multivariate {S}tein factors for a class of strongly log-concave
  distributions.
\newblock \emph{Electron. Commun. Probab.}, 21:\penalty0 14 pp., 2016.
\newblock \doi{10.1214/16-ECP15}.

\bibitem[M{\"u}ller(1997)]{Muller97}
M{\"u}ller, A.
\newblock Integral probability metrics and their generating classes of
  functions.
\newblock \emph{Ann. Appl. Probab.}, 29\penalty0 (2):\penalty0 pp. 429--443,
  1997.

\bibitem[Oates \& Girolami(2015)Oates and Girolami]{Oates2015}
Oates, C. and Girolami, M.
\newblock Control functionals for {Q}uasi-{M}onte {C}arlo integration.
\newblock \emph{arXiv:1501.03379}, 2015.

\bibitem[Oates et~al.(2016{\natexlab{a}})Oates, Cockayne, Briol, and
  Girolami]{Oates2016}
Oates, C., Cockayne, J., Briol, F., and Girolami, M.
\newblock Convergence rates for a class of estimators based on stein’s
  method.
\newblock \emph{arXiv preprint arXiv:1603.03220}, 2016{\natexlab{a}}.

\bibitem[Oates et~al.(2016{\natexlab{b}})Oates, Girolami, and
  Chopin]{OatesGiCh2016}
Oates, C.~J., Girolami, M., and Chopin, N.
\newblock Control functionals for {M}onte {C}arlo integration.
\newblock \emph{Journal of the Royal Statistical Society: Series B (Statistical
  Methodology)}, pp.\  n/a--n/a, 2016{\natexlab{b}}.
\newblock ISSN 1467-9868.
\newblock \doi{10.1111/rssb.12185}.

\bibitem[Ranganath et~al.(2016)Ranganath, Tran, Altosaar, and
  Blei]{Ranganath2016}
Ranganath, R., Tran, D., Altosaar, J., and Blei, D.
\newblock Operator variational inference.
\newblock In \emph{Advances in Neural Information Processing Systems}, pp.\
  496--504, 2016.

\bibitem[Simon-Gabriel \& Sch{\"o}lkopf(2016)Simon-Gabriel and
  Sch{\"o}lkopf]{Simon2016kernel}
Simon-Gabriel, C. and Sch{\"o}lkopf, B.
\newblock Kernel distribution embeddings: Universal kernels, characteristic
  kernels and kernel metrics on distributions.
\newblock \emph{arXiv preprint arXiv:1604.05251}, 2016.

\bibitem[Sriperumbudur(2016)]{sriperumbudur2016optimal}
Sriperumbudur, B.
\newblock On the optimal estimation of probability measures in weak and strong
  topologies.
\newblock \emph{Bernoulli}, 22\penalty0 (3):\penalty0 1839--1893, 2016.

\bibitem[Sriperumbudur et~al.(2010)Sriperumbudur, Gretton, Fukumizu,
  Sch{\"o}lkopf, and Lanckriet]{sriperumbudur2010hilbert}
Sriperumbudur, B., Gretton, A., Fukumizu, K., Sch{\"o}lkopf, B., and Lanckriet,
  G.
\newblock Hilbert space embeddings and metrics on probability measures.
\newblock \emph{J. Mach. Learn. Res.}, 11\penalty0 (Apr):\penalty0 1517--1561,
  2010.

\bibitem[Stein(1972)]{Stein72}
Stein, C.
\newblock A bound for the error in the normal approximation to the distribution
  of a sum of dependent random variables.
\newblock In \emph{Proc. 6th {B}erkeley {S}ymposium on {M}athematical
  {S}tatistics and {P}robability ({U}niv. {C}alifornia, {B}erkeley, {C}alif.,
  1970/1971), {V}ol. {II}: {P}robability theory}, pp.\  583--602. Univ.
  California Press, Berkeley, Calif., 1972.

\bibitem[Stein et~al.(2004)Stein, Diaconis, Holmes, and Reinert]{SteinDiHoRe04}
Stein, C., Diaconis, P., Holmes, S., and Reinert, G.
\newblock Use of exchangeable pairs in the analysis of simulations.
\newblock In \emph{Stein's method: expository lectures and applications},
  volume~46 of \emph{IMS Lecture Notes Monogr. Ser.}, pp.\  1--26. Inst. Math.
  Statist., Beachwood, OH, 2004.

\bibitem[Steinwart \& Christmann(2008)Steinwart and Christmann]{Christmann2008}
Steinwart, I. and Christmann, A.
\newblock \emph{Support Vector Machines}.
\newblock Springer Science \& Business Media, 2008.

\bibitem[Stewart(1976)]{Stewart1976}
Stewart, J.
\newblock Positive definite functions and generalizations, an historical
  survey.
\newblock \emph{Rocky Mountain J. Math.}, 6\penalty0 (3):\penalty0 409--434, 09
  1976.
\newblock \doi{10.1216/RMJ-1976-6-3-409}.

\bibitem[Vallender(1974)]{Vallender74}
Vallender, S.
\newblock Calculation of the {W}asserstein distance between probability
  distributions on the line.
\newblock \emph{Theory Probab. Appl.}, 18\penalty0 (4):\penalty0 784--786,
  1974.

\bibitem[Wainwright(2017)]{Wainwright17}
Wainwright, M.
\newblock \emph{High-dimensional statistics: A non-asymptotic viewpoint}.
\newblock 2017.
\newblock URL
  \url{http://www.stat.berkeley.edu/~wainwrig/nachdiplom/Chap5_Sep10_2015.pdf}.

\bibitem[{Wang} \& {Liu}(2016){Wang} and {Liu}]{WangLi2016}
{Wang}, D. and {Liu}, Q.
\newblock {Learning to Draw Samples: With Application to Amortized MLE for
  Generative Adversarial Learning}.
\newblock \emph{arXiv:1611.01722}, November 2016.

\bibitem[Welling \& Teh(2011)Welling and Teh]{WellingTe11}
Welling, M. and Teh, Y.
\newblock Bayesian learning via stochastic gradient {L}angevin dynamics.
\newblock In \emph{ICML}, 2011.

\bibitem[Wendland(2004)]{Wendland2004}
Wendland, H.
\newblock \emph{Scattered data approximation}, volume~17.
\newblock Cambridge university press, 2004.

\bibitem[Zellner \& Min(1995)Zellner and Min]{ZellnerMi95}
Zellner, A. and Min, C.
\newblock Gibbs sampler convergence criteria.
\newblock \emph{JASA}, 90\penalty0 (431):\penalty0 921--927, 1995.

\end{thebibliography}
\bibliographystyle{icml2017}

\onecolumn
\appendix

\section{Additional appendix notation}

We use $f \ast h$ to denote the convolution between $f$ and $h$, and, for absolutely integrable $f: \reals^d \to \reals$, we say 
$\hat{f}(\omega) \defeq (2\pi)^{-d/2} \int f(x) e^{-i \inner{x}{\omega}} dx$ is the Fourier transform of $f$.
For $g \in \kset_k^d$ we define $\kdnorm{g} \defeq
\sqrt{\sum_{j=1}^d \knorm{g_j}^2}$.  
Let $L^2$ denote the Banach space of real-valued functions $f$ with
$\norm{f}_{L^2}\defeq \int f(x)^2 \dx < \infty$. 
For $\reals^d$-valued $g$, we will
overload $g\in L^2$ to mean
$\norm{g}_{L^2} \defeq \sqrt{\sum_{j=1}^d \norm{g_j}_{L^2}^2} < \infty$.
We define the operator norm of a vector $a\in\reals^d$ as
$\opnorm{a}\defeq \twonorm{a}$ and of a matrix $A\in\reals^{d\times d}$
as $\opnorm{A} \defeq \sup_{x\in\reals^d,\twonorm{x}=1} \twonorm{Ax}$. 
We further define the Lipschitz constant $M_2(g)\defeq \sup_{x\neq y} \opnorm{\grad g(x) - \grad
g(y)}/\twonorm{x-y}$ and the ball $\ball(x, r) \defeq
\{y\in\reals^d\,|\,\twonorm{x-y}\leq r\}$ for any $x\in\reals^d$ and $r \geq 0$.

\section{Proof of \propreflow{stein-class}: Zero mean test functions}\label{sec:stein-class-proof}

Fix any $g \in \ksteinset{}$.
Since $k\in C^{(1,1)}$, $\sup_{x\in\reals^d} k(x,x) < \infty$, and
$\sup_{x\in\reals^d} \opnorm{\grad_x \grad_y k(x,x)} < \infty$, Cor. 4.36 of 
\citep{Christmann2008} implies that
$M_0(g_j)  < \infty$ and
$M_1(g_j)  < \infty$
for each $j \in \{1,\dots, d\}$.
As $\Esubarg{P}{\twonorm{b(Z)}} < \infty$, the proof of \citep[Prop. 1]{GorhamMa15} now implies
$\Esubarg{P}{\langarg{g}{Z}} = 0$.

\section{Proof of \propreflow{kernel-stein-discrepancy-formula}: KSD closed form}\label{sec:kernel-stein-discrepancy-formula-proof}
Our proof generalizes that of \citep[Thm.\ 2.1]{ChwialkowskiStGr2016}.
For each dimension $j\in\{1,\dots, d\}$, we define the operator
$\langcomp{j}{}$ via 
$(\langcomp{j}{g_0})(x) 
	\defeq \frac{1}{p(x)}\grad_{x_j}(p(x) g_0(x))
	= \grad_{x_j}g_0(x) +b_j(x) g_0(x)$ for $g_0:\reals^d\to\reals$.  
We further let $\Psi_k:\reals^d\to\kset_k$ denote the canonical feature map of $\kset_k$, given by
$\Psi_k(x)\defeq k(x,\cdot)$. 
Since $k \in C^{(1,1)}$, the argument of \citep[Cor. 4.36]{Christmann2008}
implies that
\balign\label{eqn:langevin-componentwise}
  \langevin{g}(x)
    &\textstyle= \sum_{j=1}^d (\langcomp{j}{g_j})(x)
    = \sum_{j=1}^d \langcomp{j}{\inner{g_j}{\Psi_k(x)}_{\kset_k}}
    = \sum_{j=1}^d \inner{g_j}{\langcomp{j}{}\Psi_k(x)}_{\kset_k}
\ealign
for all $g=(g_1,\dots,g_d)\in\ksteinsetnorm{k}{\norm{\cdot}}$ and $x\in\reals^d$.
Moreover, \citep[Lem. 4.34]{Christmann2008} gives
\balign\label{eqn:kj-derivation}
\inner{\langcomp{j}\Psi_k(x)}{\langcomp{j}\Psi_k(y)}
    &= \langle b_j(x)\Psi_k(x) + \grad_{x_j}\Psi_k(x), b_j(y)\Psi_k(y) + \grad_{y_j}\Psi_k(y)\rangle_{\kset_k} \notag \\
    &= b_j(x)b_j(y)k(x,y) + b_j(x)\grad_{y_j}k(x,y) + b_j(y)\grad_{x_j}k(x,y) + \grad_{x_j}\grad_{y_j}k(x,y)
    = k_0^j(x,y)
\ealign
for all $x,y\in\reals^d$ and $j\in\{1,\dots, d\}$.
The representation \eqnref{kj-derivation} and our $\mu$-integrability assumption together imply
that, for each $j$, $\langcomp{j}{\Psi_k}$ is Bochner $\mu$-integrable \citep[Definition A.5.20]{Christmann2008}, since
\baligns
\Esubarg{\mu}{\knorm{\langcomp{j}{\Psi_k(X)}}}
  = \Esubarg{\mu}{\sqrt{k_0^j(X, X)}} < \infty.
\ealigns
Hence, we may apply the representation \eqnref{kj-derivation} and exchange expectation and RKHS inner product to discover
\balign\label{eqn:kj-closed-form}
w_j^2
   = \Earg{k_0^j(X, \tilde{X})}
   = \Earg{\inner{\langcomp{j}{\Psi_k(X)}}
    {\langcomp{j}{\Psi_k(\tilde{X})}}_{\kset_k}}
   = \knorm{\Esubarg{\mu}{\langcomp{j}{\Psi_k(\QVAR)}}}^2.
\ealign
for $X, \tilde{X} \iid \mu$.
To conclude, we invoke the representation \eqnref{langevin-componentwise}, Bochner $\mu$-integrability, the representation \eqnref{kj-closed-form}, and the Fenchel-Young inequality for dual norms twice:
\baligns
\langstein{\mu}{\ksteinsetnorm{k}{\norm{\cdot}}}
  &= \sup_{g\in\ksteinsetnorm{k}{\norm{\cdot}}} \Esub{\mu}[\langarg{g}{\QVAR}]
  = \sup_{\knorm{g_j} = v_j, \norm{v}^* \leq 1} \textsum_{j=1}^d
     \inner{g_j}{\Esub{\mu}[\langcomp{j}{\Psi_k(\QVAR)}]}_{\kset_k} \\
  &= \sup_{\norm{v}^* \leq 1} \textsum_{j=1}^d v_j
    \knorm{\Esub{\mu}[\langcomp{j}{\Psi_k(\QVAR)}]}
  = \sup_{\norm{v}^*\le 1} \textsum_{j=1}^d v_j w_j
  = \norm{w}.
\ealigns

\section{Proof of \propreflow{stein-set-equivalence}: Stein set equivalence}\label{sec:stein-set-equivalence-proof}
By \propref{kernel-stein-discrepancy-formula}, $\langstein{\mu}{\ksteinsetnorm{k}{\norm{\cdot}}} =
\norm{w}$ and $\langstein{\mu}{\ksteinsetnorm{k}{\twonorm{\cdot}}} =
\twonorm{w}$ for some vector $w$,
and by \citep[Thm. 8.7]{Bachman1966functional}, there exist
constants $c_d, c_d'> 0$ depending only on $d$ and $\norm{\cdot}$ such that
$%
c_d \norm{w} \le \twonorm{w} \le c_d '\norm{w}
$.

\section{Proof of \thmref{univariate-weak-convergence}: Univariate KSD detects non-convergence} \label{sec:univariate-weak-convergence-proof}
While the statement of \thmref{univariate-weak-convergence} applies only to
the univariate case $d=1$, we will prove all steps for general $d$ when possible.
Our strategy is to define a reference IPM $\ipm$ for which $\mu_m\Rightarrow P$ whenever $\ipm(\mu_m, P)\to 0$
and then upper bound $\ipm$ by a function of the KSD $\langstein{\mu_m}{\ksteinset{k}}$.
To construct the reference class of test functions $\hset$, we choose some \emph{integrally strictly positive definite}
(ISPD) kernel $k_b:\reals^d\times\reals^d\to\reals$, that is, we select a kernel
function $k_b$ such that
\baligns
\int_{\reals^d\times\reals^d} k_b(x,y) d\mu(x)d\mu(y) > 0
\ealigns
for all finite non-zero signed Borel measures $\mu$ on $\reals^d$
\citep[Section 1.2]{sriperumbudur2010hilbert}. 
For this proof, we will choose the Gaussian kernel $k_b(x,y) =
\exp{-\twonorm{x-y}^2/2}$, which is ISPD by \citep[Section 3.1]{sriperumbudur2010hilbert}.
Since $r(x) \defeq \exp{-\twonorm{x}^2/2}$ is bounded and continuous and never vanishes,
the kernel $\tilde{k_b}(x,y) = k_b(x,y)r(x)r(y)$ is also ISPD. 
Let $\hset \defeq \{h \in\kset_{\tilde{k_b}}\, |\, \norm{h}_{\tilde{k_b}}\le 1\}$.
By \citep[Thm. 3.2]{sriperumbudur2016optimal},
since $\tilde{k_b}$ is ISPD with $\tilde{k_b}(x,\cdot)\in C_0(\reals^d)$ for all $x$, we know
that $d_{\hset}(\mu_m, P)\to 0$ only if $\mu_m\Rightarrow P$.
With $\hset$ in hand, \thmref{univariate-weak-convergence} will follow from
our next theorem which upper bounds the IPM $\ipm(\mu, P)$ in terms of the
KSD $\langstein{\mu}{\ksteinset{k}}$.

\begin{theorem}[Univariate KSD lower bound]\label{thm:ksd-univariate-lower-bound}
Let $d=1$, and consider the set of univariate functions $\hset = \{h
\in\kset_{\tilde{k_b}}\, |\, \norm{h}_{\tilde{k_b}}\le 1\}$.
Suppose $P\in\pset$ and $k(x,y) = \Phi(x - y)$ for $\Phi \in C^2$ with
generalized Fourier transform $\hat \Phi$ and
$F(t) \defeq \sup_{\infnorm{\omega}\le t} \hat{\Phi}(\omega)^{-1}$
finite for all $t > 0$. Then there exists a
constant $\mathcal{M}_P > 0$ such that, for all probability measures $\mu$ and $\eps > 0$,
\baligns\textstyle
\ipm(\mu, P) \le \eps + \left(\frac{\pi}{2}\right)^{1/4}\mathcal{M}_P F\left
(\frac{12\log 2}{\pi}(1 + \sqrt{d-1} + M_1(b)\mathcal{M}_P)\eps^{-1}\right )^{1/2}\langstein{\mu}{\ksteinset{k}}.
\ealigns
\end{theorem}

\begin{remarks}
An explicit value for the \emph{Stein factor} $\mathcal{M}_P$ can be derived from the proof in \secref{ksd-univariate-lower-bound-proof} and the results of \citet{GorhamDuVoMa16}.
After optimizing the bound $\ipm(\mu, P)$ over $\eps > 0$, 
the Gaussian, inverse
multiquadric, and \Matern ($v > 1$) kernels achieve rates of
$O(1/\sqrt{\log(\frac{1}{\langstein{\mu}{\ksteinset{k}}})})$,
$O(1/\log(\frac{1}{\langstein{\mu}{\ksteinset{k}}}))$, and
$O(\langstein{\mu}{\ksteinset{k}}^{1/(v + 1/2)})$ respectively as
$\langstein{\mu}{\ksteinset{k}}\to 0$.
\end{remarks}

In particular, since $\hat{\Phi}$ is non-vanishing, $F(t)$ is finite for all $t$.
If $\langstein{\mu_m}{\ksteinset{k}}\to 0$, then, for any fixed $\eps > 0$, we
have $\lim\sup_{m\to \infty} \ipm(\mu_m, P) \le \eps$. Taking $\eps\to 0$ shows that
$\lim_{m\to \infty}\ipm (\mu_m,P)\to 0$, which implies that $\mu_m\Rightarrow P$.

\subsection{Proof of \thmref{ksd-univariate-lower-bound}: Univariate KSD lower bound}
\label{sec:ksd-univariate-lower-bound-proof}
Fix any probability measure $\mu$ and $h \in \hset$, 
and define the tilting function $\Xi(x)\defeq (1+\twonorm{x}^2)^{1/2}$.
The proof will proceed in three steps. 
\paragraph{Step 1: Uniform bounds on $M_0(h)$, $M_1(h)$ and
  $\sup_{x\in\reals^d} \twonorm{\Xi(x)\grad h(x)}$}
We first bound $M_0(h)$, $M_1(h)$ and $\sup_{x\in\reals^d} \twonorm{\Xi(x)\grad h(x)}$ uniformly over $\hset$.
To this end, we define the finite value
$c_0 \defeq \sup_{x\in\reals^d} (1+\twonorm{x}^2) r(x) = 2e^{-1/2}$.
For all $x\in\reals^d$, we have
\baligns
|h(x)|
  = |\inner{h}{\tilde{k_b}(x,\cdot)}_{\kset_{\tilde{k_b}}}|
  \le \knormarg{h}{\tilde{k_b}}\, \tilde{k_b}(x,x)^{1/2}
  \le 1.
\ealigns
Moreover, we have $\grad_x k_b(x,y) = (y-x)k_b(x,y)$ and $\grad r(x) = -xr(x)$. Thus
for any $x$, by \citep[Corollary 4.36]{Christmann2008} we have
\baligns
\twonorm{\grad h(x)}
  \le \knormarg{h}{\tilde{k_b}}\, \inner{\grad_x}{\grad_y \tilde{k_b}(x,x)}^{1/2}
  \le [d\,r(x)^2 + \twonorm{x}^2\,r(x)^2]^{1/2} k_b(x,x)^{1/2}
  \le [(d-1)^{1/2} + (1 + \twonorm{x}^2)^{1/2})] r(x),
\ealigns
where in the last inequality we used the triangle inequality.
Hence $\twonorm{\grad h(x)}\le (d-1)^{1/2} + 1$ and  $\twonorm{\Xi(x) \grad
  h(x)} \le (d-1)^{1/2} + c_0$ for all $x$,
completing our bounding of $M_0(h)$, $M_1(h)$ and $\sup_{x\in\reals^d}
\twonorm{\Xi(x)\grad h(x)}$ uniformly over $\hset$.

\paragraph{Step 2: Uniform bound on $\norm{g_h}_{L^2}$ for Stein solution $g_h$}
We next show that there is a solution to the $P$
\emph{Stein equation} 
\balign\label{eqn:univariate-stein-equation}
\langarg{g_h}{x} = h(x) - \Esubarg{P}{h(Z)}
\ealign
with $g_h(x) \le \mathcal{M}_P / (1 + \twonorm{x}^2)^{1/2}$ for every $h \in \hset$.
When $d=1$, this will imply that $\norm{g_h}_{L^2}$ is bounded uniformly over $\hset$. 
To proceed, we will define a tilted distribution $\tilde P \in \pset$ and a tilted function $f$, show that a solution $\tilde{g}_f$
to the $\tilde P$ Stein equation is bounded, and construct a solution $g_h$ to the Stein equation of $P$ based on $\tilde{g}_f$.

Define
$\tilde P$ via the tilted probability density 
$\tilde{p}(x) \propto p(x)/\Xi(x)$ with score function 
$\tilde{b}(x) \defeq \grad \log \tilde{p}(x) = b(x) - \xi(x)$ for $\xi(x)
\defeq \grad \log \Xi(x) = x / (1 + \twonorm{x}^2)$.
Since $b$ is Lipschitz and $\grad \xi(x) =
(1 + \twonorm{x}^2)^{-1}[I - 2 \frac{xx^{\top}}{1 + \twonorm{x}^2}]$ has its
operator norm uniformly bounded by $3$, $\tilde{b}$ is also Lipschitz.
To see that $\tilde{P}$ is also distantly dissipative, note first that
$|\inner{\xi(x) - \xi(y)}{x - y}| \le \twonorm{\xi(x) - \xi(y)}\cdot \twonorm{x-y} \le \twonorm{x - y}$
since $\sup_x \twonorm{\xi(x)} \le 1/2$. Because $P$ is distantly
dissipative, we know $\inner{b(x) - b(y)}{x-y}\le -\half\kappa_0 \twonorm{x-y}^2$ for
some $\kappa_0 > 0$ and all $\twonorm{x-y}\ge R$ for some $R > 0$. Thus for all $\twonorm{x-y}
\ge \max (R, 4/\kappa_0)$, we have
\baligns
\inner{\tilde{b}(x) - \tilde{b}(y)}{x-y}
  = \inner{b(x) - b(y)}{x-y} + \inner{\xi(x) - \xi(y)}{x-y}
  \le -\half \kappa_0 \twonorm{x-y}^2 + \twonorm{x-y} \le -\half\frac{\kappa_0}{2}\twonorm{x-y}^2,
\ealigns
so $\tilde{P}$ is also distantly dissipative and hence in $\pset$.

Let $f(x) \defeq \Xi(x) (h(x) - \Esubarg{P}{h(Z)})$.
Since $\Esubarg{\tilde P}{f(Z)} = \Esubarg{P}{h(Z) - \Esubarg{P}{h(Z)}} = 0$,
Thm. 5 and Sec. 4.2 of \cite{GorhamDuVoMa16}, imply that 
the $\tilde{P}$ Stein equation $(\mathcal{T}_{\tilde P}{\tilde g_f})(x) = f(x)$ 
has a solution $\tilde g_f$ with $M_0(g_f) \le \mathcal{M}_P' M_1(f)$
for $\mathcal{M}_P'$ a constant independent of $f$ and $h$.
Since
$\grad f(x) = \grad\Xi(x) (h(x) - \Esubarg{P}{h(Z)}) + \Xi(x) \grad h(x)$
and $\twonorm{\grad \Xi(x)} = \frac{\twonorm{x}}{(1 + \twonorm{x}^2)^{1/2}}$ is bounded by $1$,
$M_0(g_f) \le \mathcal{M}_P'(2 + (d-1)^{1/2} + c_0) \defeq \mathcal{M}_P$, a constant
independent of $h$.

Finally, we note that $g_h(x) \defeq \tilde{g}_f(x)/\Xi(x)$ is a solution to
the $P$ Stein equation \eqnref{univariate-stein-equation} satisfying $g_h(x)
\le \mathcal{M}_P / \Xi(x) = \mathcal{M}_P / (1 + \twonorm{x}^2)^{1/2}$.
Hence, in the case $d=1$, we have $\norm{g_h}_{L^2} \leq \mathcal{M}_P\sqrt{\pi}$.

\paragraph{Step 3: Approximate $\langevin{g_h}$ using $\langevin{\ksteinset{k}}$}
In our final step, we will use the following lemma, proved in \secref{stein-solution-l2-has-finite-rkhs-norm-proof}, to show that we can
approximate $\langevin{g_h}$ arbitrarily well by a function in a scaled copy
of $\langevin{\ksteinset{k}}$. 

\begin{lemma}[Stein approximations with finite RKHS norm]\label{lem:stein-solution-l2-has-finite-rkhs-norm}
Suppose that $g:\reals^d\to\reals^d$ is bounded and belongs to
$L^2\cap C^1$ and that $h = \langevin{g}$ and $\grad \log p$ are Lipschitz.
Moreover, suppose $k(x,y) = \Phi(x - y)$ for $\Phi \in C^2$ with
generalized Fourier transform $\hat \Phi$. Then for every $\eps > 0$, there is
a function $g_{\eps}:\reals^d\to\reals^d$ such that $\sup_{x\in\reals^d}
|\langarg{g_{\eps}}{x} - \langarg{g}{x}|\le \eps$ and
\baligns
\textstyle
\kdnorm{g_{\eps}} \le (2\pi)^{-d/4} F\left (\frac{12d \log
  2}{\pi} (M_1(h) + M_1(b)M_0(g))\eps^{-1}\right )^{1/2} \norm{g}_{L^2},
\ealigns
where $F(t) \defeq \sup_{\infnorm{\omega}\le t} \hat{\Phi}(\omega)^{-1}$.
\end{lemma}
When $d=1$, \lemref{stein-solution-l2-has-finite-rkhs-norm} implies that for every
$\eps > 0$ there is a function $g_{\eps}:\reals\to\reals$ such that
$M_0(\langevin{g_{\eps}}- h)\le \eps$  and
$\knorm{g_{\eps}}\le (\frac{\pi}{2})^{1/4} \mathcal{M}_P
F(\frac{12\log 2}{\pi} (M_1(h) + M_1(b)\mathcal{M}_P)\eps^{-1})^{1/2}$. Hence we have
\baligns
|\Esubarg{P}{h(\PVAR)} - \Esubarg{\mu}{h(\QVAR)}|
  &\le |\Esubarg{\mu}{h(\QVAR) - \langarg{g_{\eps}}{\QVAR}}| +
  |\Esubarg{\mu}{\langarg{g_{\eps}}{\QVAR}}| \\
  &\le \eps + \knorm{g_{\eps}} \langstein{\mu}{\ksteinset{k}} \\
  &\textstyle\le \eps + (2\pi)^{-1/4} \mathcal{M}_P \sqrt{\pi} F\left (\frac{12\log
    2}{\pi} (M_1(h) + M_1(b)\mathcal{M}_P)\eps^{-1}\right )^{1/2} \langstein{\mu}{\ksteinset{k}}.
\ealigns
Taking a supremum over $h\in\hset$ yields the advertised result.

\subsection{Proof of \lemref{stein-solution-l2-has-finite-rkhs-norm}: Stein
  approximations with finite RKHS norm} \label{sec:stein-solution-l2-has-finite-rkhs-norm-proof}

Let us define the function $S:\reals^d\to\reals$ via the mapping $S(x) \defeq
\prod_{j=1}^d \frac{\sin x_j}{x_j}$. Then $S\in L^2$ and $\int_{\reals^d}
\twonorm{x}S(x)^4<\infty$. We will then define the
density function $\rho(x) \defeq \mathcal{Z}^{-1} S(x)^4$, where $\mathcal{Z} \defeq
\int_{\reals^d} S(x)^4\dx = (2\pi/3)^{d}$ is the normalization constant.
One can check that $\hat{\rho}(\omega)^2 \le (2\pi)^{-d}\indic{\infnorm{\omega}\le 4}$.

Let $Y$ be a random variable with density
$\rho$. For each $\delta > 0$, let us define $\rho_{\delta}(x) =
\delta^{-d}\rho(x/\delta)$ and for any function $f$ let us denote
$f_{\delta}(x) \defeq \Earg{f(x - \delta Y)}$.
Since $h = \langevin{g}$ is assumed Lipschitz, this implies
$|h_{\delta}(x) - h(x)| = |\Esubarg{\rho}{h(x - \delta Y) - h(x)}| \le
\delta\, M_1(h)\, \Esubarg{\rho}{\twonorm{Y}}$ for all $x\in\reals^d$.

Next, notice that for any $\delta > 0$ and $x\in\reals^d$,
\baligns
\langarg{g_{\delta}}{x} &= \Esubarg{\rho}{\inner{b(x)}{g(x - \delta Y)}} +
\Earg{\inner{\grad}{g(x - \delta Y)}},\qtext{and} \\
h_{\delta}(x) &= \Esubarg{\rho}{\inner{b(x - \delta Y)}{g(x - \delta Y)}} +
\Earg{\inner{\grad}{g(x - \delta Y)}}.
\ealigns
Because we assume $b$ is Lipschitz, we can deduce from above for
any $x\in\reals^d$,
\baligns
|\langarg{g_{\delta}}{x} - h_{\delta}(x)|
  &= |\Esubarg{\rho}{\inner{b(x)- b(x - \delta Y)}{g(x - \delta Y)}}| \\
  &\le \Esubarg{\rho}{\twonorm{b(x)- b(x - \delta Y)}\twonorm{g(x - \delta Y)}} \\
  &\le M_0(g) \, M_1(b)\, \delta\, \Esubarg{\rho}{\twonorm{Y}}.
\ealigns
Thus for any $\delta > 0$, by the triangle inequality, we have
\balign\label{eqn:conv-langevin-closeness}
|\langarg{g_\delta}{x} - \langarg{g}{x}|
  \le |\langarg{g_\delta}{x} - h_{\delta}(x)| + |h_{\delta}(x) - h(x)|
  \le \delta (M_1(h) + M_1(b) M_0(g))\Esubarg{\rho}{\twonorm{Y}}.
\ealign
Letting $\tilde{\eps} = \eps/((M_1(h) +
M_1(b)M_0(g))\Esubarg{\rho}{\twonorm{Y}})$ for any $\eps > 0$, we have $M_0(\langevin{g_{\tilde{\eps}}}
- \langevin{g}) \le \eps$.

Thus it remains to bound the RKHS norm of $g_{\delta}$. By the Convolution
Theorem \citep[Thm. 5.16]{Wendland2004}, we have $\hat{g_{\delta}}(\omega) =
(2\pi)^{d/2}\hat{g}(\omega)\hat{\rho_{\delta}}(\omega)$, and so the squared norm of
$g_{\delta}$ in $\kset_k^d$ is equal to \citep[Thm. 10.21]{Wendland2004}
\baligns
(2\pi)^{-d/2}\int_{\reals^d}\frac{|\hat{g_{\delta}}(\omega)|^2}{\hat{\Phi}(\omega)}\,d\omega
  = (2\pi)^{d/2}\int_{\reals^d}\frac{|\hat{g}(\omega)|^2\hat{\rho_{\delta}}(\omega)^2}{\hat{\Phi}(\omega)}\,d\omega
  \le (2\pi)^{-d/2}\left \{\sup_{\infnorm{\omega}\le 4\delta^{-1}} \hat{\Phi}(\omega)^{-1}\right\} \int_{\reals^d}|\hat{g}(\omega)|^2 \, d\omega,
\ealigns
where in the inequality we used the fact that
$\hat{\rho_{\delta}}(\omega) = \hat{\rho}(\delta\omega)$.
By Plancherel's theorem \citep[Thm. 1.1]{herb2011plancherel}, we know
that $f\in L^2$ implies that $\norm{f}_{L^2} =\staticnorm{\hat{f}}_{L^2}$.
Thus we have $\kdnorm{g_{\delta}} \le (2\pi)^{-d/4}
F(4\delta^{-1})^{1/2} \norm{g}_{L^2}$. The final result follows from
noticing that $\int_{\reals} \sin^4(x)/x^4\dx = \frac{2\pi}{3}$ and also
\baligns
\int_{\reals^d} \twonorm{x} \prod_{j=1}^d \frac{\sin^4 x_j}{x_j^4}\dx
  \le \int_{\reals^d} \onenorm{x} \prod_{j=1}^d \frac{\sin^4 x_j}{x_j^4}\dx
  = \sum_{j=1}^d \int_{\reals^d} \frac{(\sin x_j)^4}{|x_j|^3} \prod_{k\neq
    j} \frac{\sin^4 x_k}{x_k^4}\dx
  = 2 d (\log 2)\left (\frac{2\pi}{3}\right )^{d-1},
\ealigns
which implies $\Esubarg{\rho}{\twonorm{Y}} \le \frac{3d\log 2}{\pi}$.

\section{Proof of \thmref{counterexample-theorem}: KSD fails with light kernel tails}\label{sec:counterexample-theorem-proof}
First, define the generalized inverse function
$\gamma^{-1}(s) \defeq \inf\{r\ge 0\,|\,\gamma(r)\le s\}$.
Next, fix an $n\ge 1$, let $\Delta_n \defeq \max(1,\gamma^{-1}(1/n))$, and define $r_n \defeq \Delta_n n^{1/d}$.
Select $n$ distinct points
$x_1, \dots, x_n\in\reals^d$ so that $z_{i,i'}\defeq x_i - x_{i'}$ satisfies
$\twonorm{z_{i,i'}} > \Delta_n$ for all $i\neq i'$ and $\twonorm{x_i} \leq r_n$ for all $i$.
By \citep[Lems. 5.1 and 5.2]{Wainwright17}, such a point set always exists.
Now define $Q_n = \frac{1}{n}\sum_{i=1}^n \delta_{x_i}$.
We will show that if $\Delta_n$ grows at an appropriate rate then $\langstein{Q_n}{\ksteinset{k}}\to 0$ as $n\to\infty$.

Since the target distribution $P$ is $\Gsn(0,I_d)$, the associated gradient
of the log density is $b(x) = -x$. Thus
\baligns k_0(x,y)
  \defeq \sum_{j=1}^d k_0^j(x,y)
  = \inner{x}{y}k(x,y) -
    \inner{y}{\grad_x k(x,y)} -
    \inner{x}{\grad_y k(x,y)} +
    \inner{\grad_x}{\grad_y k(x,y)}.
\ealigns
From \propref{kernel-stein-discrepancy-formula}, we have
\balign\label{eqn:ksd-counterexample-decomp}
\langstein{Q_n}{\ksteinset{k}}^2
  = \frac{1}{n^2}\sum_{i,i'=1}^n k_0(x_i, x_{i'})
  = \frac{1}{n^2}\sum_{i=1}^n k_0(x_i, x_i) + \frac{1}{n^2}\sum_{i\neq i'} k_0(x_i, x_{i'}).
\ealign
Since $k \in C^{(2,2)}_b$, $\gamma(0) < \infty$.
Thus by Cauchy-Schwarz, the first term of \eqnref{ksd-counterexample-decomp} is
upper bounded by
\baligns
\frac{1}{n^2}\sum_{i=1}^n k_0(x_i, x_i)
  &\le \frac{1}{n^2}\sum_{i=1}^n \twonorm{x_i}^2 k(x_i, x_i) +
  \twonorm{x_i} (\twonorm{\grad_x k(x_i, x_i)} + \twonorm{\grad_y k(x_i,
    x_i)}) +
  |\inner{\grad_x}{\grad_y k(x_i, x_i)}| \\
  &\le \frac{\gamma(0)}{n} [r_n^2 + 2r_n + 1] \leq \frac{\gamma(0)}{n} (n^{1/d}\Delta_n + 1)^2.
\ealigns
To handle the second term of \eqnref{ksd-counterexample-decomp},
we will use the assumed bound on $k$ and its derivatives from $\gamma$.
For any fixed $i\neq i'$, by the
triangle inequality, Cauchy-Schwarz, and fact $\gamma$ is monotonically
decreasing we have
\baligns%
|k_0(x_i, x_{i'})|
&\le \twonorm{x_i}\twonorm{x_{i'}} |k(x_i, x_{i'})|
  + \twonorm{x_i}\twonorm{\grad_y k(x_i, x_{i'})}
  + \twonorm{x_{i'}}\twonorm{\grad_x k(x_i, x_{i'})}
  + |\inner{\grad_x}{\grad_y k(x_i,x_{i'})} | \notag \\
  &\le r_n^2 \gamma(\twonorm{z_{i,i'}})
  + r_n \gamma(\twonorm{z_{i,i'}})
  + r_n \gamma(\twonorm{z_{i,i'}})
  + \gamma(\twonorm{z_{i,i'}}) \notag \\
  &\le (n^{1/d}\Delta_n + 1)^2 \gamma(\Delta_n).
\ealigns
Our upper bounds
on the Stein discrepancy \eqnref{ksd-counterexample-decomp} and our choice of $\Delta_n$ now imply that
\[
\langstein{Q_n}{\ksteinset{k}} = O(n^{1/d - 1/2}\gamma^{-1}(1/n) + n^{-1/2}).
\]
Moreover, since $\gamma(r) = o(r^{-\alpha})$, we have $\gamma^{-1}(1/n) = o(n^{1/\alpha})=o(n^{1/2-1/d})$,
and hence $\langstein{Q_n}{\ksteinset{k}} \to 0$ as $n \to \infty$.

However, the sequence $(Q_n)_{n\geq 1}$ is not uniformly tight and hence converges to no probability measure.
This follows as, for each $r>0$,
\[
Q_m(\twonorm{X} \leq r) \leq \frac{(r+4r/{\Delta_m})^d}{m} \leq \frac{5^dr^d}{m} \leq \frac{1}{5}
\]
for $m = \lceil{5^{d+1}r^d\rceil}$, since at most $(r+4r/{\Delta_m})^d$ points with minimum pairwise Euclidean distance greater than $\Delta_m$
can fit into a ball of radius $r$~\citep[Lems. 5.1 and 5.2]{Wainwright17}.

\section{Proof of \thmref{tightness-density-implies-weak-convergence}: KSD detects tight non-convergence}\label{sec:tightness-density-implies-weak-convergence-proof}

For any probability measure $\mu$ on $\reals^d$ and $\eps > 0$, we define
its \emph{tightness rate} as
\balign\label{eqn:tightness-rate}
R(\mu, \eps) \defeq \inf \{r \geq 0 \, |\, \mu(\twonorm{\QVAR} > r) \le \eps\}. 
\ealign
\thmref{tightness-density-implies-weak-convergence}
will follow from the following result which upper bounds
the bounded Lipschitz metric $\twobl(\mu,P)$ in terms of the tightness rate
$R(\mu,\eps)$, the rate of decay of the generalized Fourier transform
$\hat{\Phi}$, and the KSD $\langstein{\mu}{\ksteinset{k}}$.

\begin{theorem}[KSD tightness lower bound]\label{thm:tightness-density-blset-upper-bound}
Suppose $P\in\pset$ and let $\mu$ be a probability measure with tightness
rate $R(\mu, \eps)$ defined in \eqnref{tightness-rate}. Moreover, suppose the kernel
$k(x,y) = \Phi(x-y)$ with $\Phi\in C^2$ and $F(t) \defeq
\sup_{\infnorm{\omega}\le t} \hat{\Phi}(\omega)^{-1}$ finite for all $t >
0$.  Then there exists a constant $\mathcal{M}_P$ such that, for all
$\rho,\eps,\delta > 0$,
\baligns
\twobl(\mu, P) &\le 
\rho \sqrt{d} (1 + M_1(b)\mathcal{M}_P) +
\eps +
  \min (\eps,1)(2+\eps+ (M_1(b) \rho \sqrt{d} + \textfrac{\delta^{-1} d \theta_{d-1}}{\theta_d} )\mathcal{M}_P) \\
&\quad + (2\pi)^{-d/4} V_d^{1/2} \mathcal{M}_P (R(\mu, \eps) +
  2\delta)^{d/2} F\left (\textfrac{12d\log 2}{\pi} (c_{\rho,\delta} +
  M_1(b)\mathcal{M}_P)\eps^{-1}\right )^{1/2} \langstein{\mu}{\ksteinset{k}},
\ealigns
where $\theta_d\defeq d \int_0^1 \exp{-1 / (1 - r^2)} r^{d-1}\dr$ for
$d > 0$ (and $\theta_0 \defeq e^{-1}$),
 $V_d$ is the volume of the unit Euclidean ball in dimension $d$, and
 \baligns
 c_{\rho,\delta} \defeq 1 + M_1(b)\mathcal{M}_P(1+d) + (2 + (M_1(b) \rho + \textfrac{1}{\rho}) \sqrt{d} \mathcal{M}_P) \textfrac{\delta^{-1} d \theta_{d-1}}{\theta_d}  + \delta^{-2} \textfrac{22}{\theta_d}\mathcal{M}_P.
 \ealigns
\end{theorem}

\begin{remarks}
An explicit value for the \emph{Stein factor} $\mathcal{M}_P$
can be derived from the proof in \secref{tightness-density-blset-upper-bound-proof} and
the results of \citet{GorhamDuVoMa16}.
When bounds on $R$ and $F$ are known, the final expression can be optimized
over $\eps, \rho$ and $\delta$ to produce
rates of convergence in $\twobl$.
\end{remarks}

Fix any $\delta > 0$, and consider a sequence of probability measures $(\mu_m)_{m\ge 1}$ that is
uniformly tight. We must have $\lim\sup_m R(\mu_m, \eps) < \infty$ for
all $\eps > 0$. Moreover, since $\hat{\Phi}$ is non-vanishing, $F(t)$ is
finite for all $t$. Thus if $\langstein{\mu_m}{\ksteinset{k}}\to 0$, then
for any fixed $\eps < 1$ and $\rho > 0$, 
\baligns
\lim\sup_m \twobl(\mu_m, P) \le 
\rho \sqrt{d} (1 + M_1(b)\mathcal{M}_P)  + \eps (3 + \eps + 
 (M_1(b) \rho \sqrt{d} + \textfrac{\delta^{-1} d \theta_{d-1}}{\theta_d} )\mathcal{M}_P).
 \ealigns  
 Taking $\rho,\eps\to 0$ yields $\twobl(\mu_m,P)\to 0$.

\subsection{Proof of \thmref{tightness-density-blset-upper-bound}: KSD tightness lower bound}
\label{sec:tightness-density-blset-upper-bound-proof}
Fix any $h\in\twoblset$. By Theorem 5 and Section 4.2 of
\cite{GorhamDuVoMa16}, there exists a $g\in C^1$ which solves the Stein
equation $\langevin{g} = h - \Earg{h(\PVAR)}$ and satisfies $M_0(g) \le
\mathcal{M}_P$ for $\mathcal{M}_P$ a constant independent of $h$ and $g$.
To show that we can approximate $\langevin{g}$ arbitrarily well by a
function in a scaled copy of $\langevin{\ksteinset{k}}$, we will make two
modifications to each $g$: first, we will approximate $g$ by a smoothened function
$g_\rho$ with $M_1(g_\rho)$ 
uniformly bounded, and second, we we truncate $g_\rho$ so that the result lies in $L^2$.

\paragraph{Smoothing $g$ by convolution}
Fix any $\rho > 0$, and 
define $g_\rho(x) \defeq \Earg{g(x - \rho U)}$ for
$U \sim \Gsn(0, I_d)$ a $d$-dimensional standard multivariate Gaussian vector.
We have $M_0(g_\rho) \le M_0(g) \leq \mathcal{M}_P$ and, invoking integration by parts,
\baligns
M_1(g_\rho) 
	&= \sup_{x\in\reals^d} \twonorm{\Earg{\grad_x g(x - \rho U)}} 
	= \sup_{x\in\reals^d} \twonorm{\Earg{g(x - \rho U) U}} /\rho 
	\le M_0(g) \Earg{\twonorm{U}}/\rho
	\leq \mathcal{M}_P \sqrt{d}/\rho.
\ealigns
Moreover, by the argument \eqnref{conv-langevin-closeness} employed in the proof of \lemref{stein-solution-l2-has-finite-rkhs-norm}, 
we have 
\baligns
M_0(\langevin{g_\rho} - \langevin{g}) \le \rho (M_1(\langevin{g}) + M_1(b)\mathcal{M}_P)
\Earg{\twonorm{U}} \leq \rho \sqrt{d} (1 + M_1(b)\mathcal{M}_P),
\ealigns
showing that $\langevin{g_\rho}$ closely approximates $\langevin{g}$.

Finally, we show that $M_0(\langevin{g_\rho})$ and $M_1(\langevin{g_\rho})$ are bounded uniformly in $g$.
Indeed, letting $h_{\rho}(x) \defeq \Earg{h(x - \rho U)}$, we see that 
$(\langevin{g_\rho})(x) = h_\rho(x) - Ph + \Earg{\inner{b(x)-b(x - \rho U)}{g(x-\rho U)}}$.
Therefore, 
\baligns
M_0(\langevin{g_\rho}) \leq M_0(h_\rho-Ph_\rho) + M_1(b)\rho \Earg{\twonorm{U}} M_0(g_\rho) \leq 2 + M_1(b) \rho \sqrt{d} \mathcal{M}_P,
\ealigns
and integration by parts implies that
\baligns
M_1(\langevin{g_\rho}) 
	&\leq M_1(h_\rho - Ph) + M_1(\langevin{g_\rho} - h_\rho + Ph) \\
  &\leq 1 + M_1(b)M_0(g_\rho) + \sup_{x\in\reals^d} 
  \twonorm{\Earg{(\grad_x g(x-\rho U) ) (b(x) - b(x - \rho U))
  - \grad_x \inner{b(x - \rho U)}{g(x-\rho U)}}} \\
  &= 1 + M_1(b)\mathcal{M}_P + \sup_{x\in\reals^d} 
  \twonorm{\Earg{\grad_u \inner{b(x - \rho U)-b(x)}{g(x-\rho U)}}/\rho} \\
  &= 1 + M_1(b)\mathcal{M}_P + \sup_{x\in\reals^d} 
  \twonorm{\Earg{U \inner{b(x)-b(x - \rho U)}{g(x-\rho U)}}/\rho} \\
  &\leq 1 + M_1(b)\mathcal{M}_P +  
  \Earg{\twonorm{U}^2}M_1(b) M_0(g_\rho)
  \leq 1 + M_1(b)\mathcal{M}_P(1+d).
\ealigns

\paragraph{Truncating $g_\rho$}
We will next truncate $g_\rho$ using the following lemma proved in \secref{smoothed-indicator-construction-proof}.

\begin{lemma}[Smoothed indicator function]\label{lem:smoothed-indicator-construction}
For any compact set $K\subset\reals^d$ and $\delta > 0$, define the set inflation
$K^{2\delta} \defeq
\{x\in\reals^d\,|\,\twonorm{x-y}\le 2\delta,\, \forall y \in K\}$.
There is a function $v_{K,\delta}:\reals^d\to [0,1]$ such that
\balign
&v_{K,\delta}(x) = 1 \text{ for all } x\in K \text{ and } v_{K,\delta}(x) =
0 \text{ for all } x\notin K^{2\delta}, \label{eqn:vk-bounds} \\
&\twonorm{\grad v_{K,\delta}(x)}\le \textfrac{\delta^{-1} d \theta_{d-1}}{\theta_d} \indic{x\in K^{2\delta}\setminus
  K} \label{eqn:grad-vk-bounds}, \\
&\opnorm{\Hess v_{K,\delta}(x)}\le \delta^{-2} \textfrac{22}{\theta_d} \indic{x\in
K^{2\delta}\setminus K} \label{eqn:hess-vk-bounds}
\ealign
where
$\theta_d\defeq d\int_0^1 \exp{-1 / (1 - r^2)} r^{d-1}\dr$ for $d > 0$
and $\theta_0 \defeq e^{-1}$.
\end{lemma}
Fix any $\eps, \delta > 0$, and let $K = \ball(0, R(\mu, \eps))$ with
$R(\mu,\eps)$ defined in \eqnref{tightness-rate}.
This set is compact since our sequence is uniformly tight.
Hence, we may define $g_{K,\delta}(x) \defeq g_\rho(x)\, v_{K,\delta}(x)$ as a smooth, truncated version of $g_\rho$ based on \lemref{smoothed-indicator-construction}.  
Since
\baligns
\langarg{g_\rho}{x} - \langarg{g_{K,\delta}}{x}
  &= (1 - v_{K,\delta}(x))[\inner{b(x)}{g_\rho(x)} + \inner{\grad}{g_\rho}(x)] + \inner{\grad
    v_{K,\delta}(x)}{g_\rho(x)} \\
  &= (1 - v_{K,\delta}(x))\langarg{g_\rho}{x} + \inner{\grad v_{K,\delta}(x)}{g_\rho(x)},
\ealigns
properties \eqnref{vk-bounds} and \eqnref{grad-vk-bounds} imply that
$\langarg{g_\rho}{x} = \langarg{g_{K,\delta}}{x}$ for all $x\in K$,
$\langarg{g_{K,\delta}}{x} = 0$ when $x\notin K^{2\delta}$, and
\baligns
|\langarg{g_\rho}{x} - \langarg{g_{K,\delta}}{x}|
  &\le |\langarg{g_\rho}{x}| + \twonorm{\grad v_{K,\delta}(x)}\,\twonorm{g_\rho(x)} \\
  &\le |\langarg{g_\rho}{x}| + \textfrac{\delta^{-1} d \theta_{d-1}}{\theta_d} \twonorm{g_\rho(x)}
  \le 2+ (M_1(b) \rho \sqrt{d} + \textfrac{\delta^{-1} d \theta_{d-1}}{\theta_d} )\mathcal{M}_P
\ealigns
for $x\in K^{2\delta}\setminus K$ by Cauchy-Schwarz.
In addition
\baligns
M_1(\langevin{g_{K,\delta}})
	&\leq M_1(\langevin{g_{\rho}} v_{K,\delta}) + M_1(\inner{\grad v_{K,\delta}}{g_\rho}) \\
	&\leq M_1(\langevin{g_{\rho}}) + M_0(\langevin{g_{\rho}})M_1(v_{K,\delta}) + M_1(\grad v_{K,\delta})M_0(g_\rho)
	+ M_0(\grad v_{K,\delta})M_1(g_\rho) \\
	&\leq 1 + M_1(b)\mathcal{M}_P(1+d) + (2 + M_1(b) \rho \sqrt{d} \mathcal{M}_P) \textfrac{\delta^{-1} d \theta_{d-1}}{\theta_d}  + \delta^{-2} \textfrac{22}{\theta_d}\mathcal{M}_P
	+ \textfrac{\delta^{-1} d \theta_{d-1}}{\theta_d} \mathcal{M}_P \sqrt{d}/\rho
	= c_{\rho,\delta}.
\ealigns
Moreover, since $v_{K,\delta}$ has compact support and is in $C^1$ by
\eqnref{vk-bounds}, $g_{K,\delta}\in C^1$
with $\norm{g_{K,\delta}}_{L^2} \leq \Vol{K^{2\delta}}^{1/2} M_0(g_\rho) \leq \Vol{K^{2\delta}}^{1/2} \mathcal{M}_P$.
Therefore, \lemref{stein-solution-l2-has-finite-rkhs-norm} implies that 
there is a function $\tilde{g}_{\eps}\in \kset_k^d$ such that
$|\langarg{\tilde{g}_{\eps}}{x} - \langarg{g_{K,\delta}}{x}| \le \eps$ for all $x$ with norm
\balign\label{eqn:geps-norm-bound}
\kdnorm{\tilde{g}_{\eps}}
  &\le (2\pi)^{-d/4} F(\textfrac{12d\log
  2}{\pi}(c_{\rho,\delta} + M_1(b) \mathcal{M}_P \eps^{-1}))^{1/2} \Vol{K^{2\delta}}^{1/2} \mathcal{M}_P.
\ealign
Using the fact that $\langevin{g_{K,\delta}}$ and $\langevin{g_\rho}$ are identical on
$K$, we have $|\langarg{\tilde{g}_{\eps}}{x} - \langarg{g_\rho}{x}|\le \eps$ for all
$x\in K$. Moreover, when $x\notin K$, the triangle inequality gives
\[
|\langarg{\tilde{g}_{\eps}}{x} - \langarg{g_\rho}{x}|
	\leq |\langarg{\tilde{g}_{\eps}}{x} - \langarg{g_{K,\delta}}{x}| + |\langarg{g_{K,\delta}}{x} - \langarg{g_\rho}{x}|
	\le 2 + \eps + (M_1(b) \rho \sqrt{d} + \textfrac{\delta^{-1} d \theta_{d-1}}{\theta_d} )\mathcal{M}_P.
\]
By the triangle inequality and the fact that our choice of $K$ ensures
$\mu(\indic{X\notin K}) \le \min(\eps, 1)$, we have
\baligns
|&\Esubarg{\mu}{h(\QVAR)} - \Esubarg{P}{h(\PVAR)}|
  = |\Esubarg{\mu}{\langarg{g}{\QVAR}}| \\
  &\le |\Earg{\langarg{g}{\QVAR} - \langarg{g_\rho}{\QVAR}}| 
  + |\Earg{\langarg{g_\rho}{\QVAR} - \langarg{\tilde{g}_{\eps}}{\QVAR}}| 
  + |\Esubarg{\mu}{\langarg{\tilde{g}_{\eps}}{\QVAR}}| \\
  &\le M_0(\langevin{g} - \langevin{g_\rho}) + |\Esubarg{\mu}{(\langarg{g_\rho}{\QVAR} - \langarg{\tilde{g}_{\eps}}{\QVAR})\indic{\QVAR\in K}}|
    + |\Esubarg{\mu}{(\langarg{g_\rho}{\QVAR} -
      \langarg{\tilde{g}_{\eps}}{\QVAR})\indic{\QVAR\notin K}}| \\
  &\qquad + |\Esubarg{\mu}{\langarg{\tilde{g}_{\eps}}{\QVAR}}| \\
  &\textstyle\le 
  \rho \sqrt{d} (1 + M_1(b)\mathcal{M}_P) 
  + \eps
  + \min(\eps,1) (2 + \eps +  (M_1(b) \rho \sqrt{d} + \textfrac{\delta^{-1} d \theta_{d-1}}{\theta_d} )\mathcal{M}_P) + \kdnorm{\tilde{g}_{\eps}}\,
  \langstein{\mu}{\ksteinset{k}} \\
    &\textstyle\le \rho \sqrt{d} (1 + M_1(b)\mathcal{M}_P) 
    + \eps + \min(\eps,1) (2 + \eps +  (M_1(b) \rho \sqrt{d} + \textfrac{\delta^{-1} d \theta_{d-1}}{\theta_d} )\mathcal{M}_P) \\
    &\textstyle\quad+ (2\pi)^{-d/4} \Vol{\ball(0, R(\mu,
    \eps) + 2\delta)}^{1/2} F\left (
    \frac{12d\log 2}{\pi} (c_{\rho,\delta} + M_1(b)\mathcal{M}_P) \eps^{-1}\right )^{1/2} \mathcal{M}_P \langstein{\mu}{\ksteinset{k}}.
\ealigns
The advertised result follows by substituting $\Vol{\ball(0, r)} = V_d r^d$ and taking the supremum over all
$h\in\blset$.

\subsection{Proof of \lemref{smoothed-indicator-construction}: Smoothed
  indicator function}
\label{sec:smoothed-indicator-construction-proof}

For all $x\in\reals^d$, define the standard normalized bump function $\psi\in
C^{\infty}$ as
\baligns
\psi(x) \defeq I_d^{-1} \exp{-1 / (1 - \twonorm{x}^2)}\indic{\twonorm{x}<1},
\ealigns
where the normalizing constant is given by
\baligns
I_d = \textstyle\int_{\ball(0,1)} \exp{-1 / (1 - \twonorm{x}^2)} \dx = \theta_d \; V_d
\ealigns
for
$V_d$ being the volume of the unit Euclidean ball in $d$
dimensions~\cite{Baker1999}.

Letting $W$ be a random variable with density $\psi$, define
$v_{K,\delta}(x)\defeq \Earg{\indic{x + \delta W \in K^{\delta}}}$ as the smoothed
approximation of $x\mapsto\indic{x\in K}$, where $\delta > 0$ controls the amount of
smoothing. Since $\supp{W} = \ball(0,1)$, we can immediately conclude
\eqnref{vk-bounds} and also $\supp{\grad v_{K,\delta}} \subseteq
K^{2\delta}\setminus K$.

Thus to prove \eqnref{grad-vk-bounds}, it remains to consider $x \in
 K^{2\delta} \setminus K$. We see $\grad v_{K,\delta}(x) =
\delta^{-d-1}\int_{\ball(x,\delta)} \grad \psi(\frac{x-y}{\delta}) \indic{y
  \in K^{\delta}} \dy$ by Leibniz rule. Letting
  $K_x^{\delta}\defeq \delta^{-1}(K^{\delta}-x)$, then by Jensen's
  inequality we have
\baligns
\twonorm{\grad v_{K,\delta}(x)}
\le \textstyle \delta^{-d-1}\int_{\ball(x,\delta)\cap K^{\delta}}
  \twonorm{\grad \psi\left (\frac{x-y}{\delta}\right )} \dy
= \delta^{-1} \int_{\ball(0,1)\cap
  K_x^{\delta}} \twonorm{\grad\psi(z)} \dz
\le \delta^{-1} \int_{\ball(0,1)}
  \twonorm{\grad\psi(z)} \dz
\ealigns
where we used the substitution $z\defeq (x-y)/\delta$. By differentiating
$\psi$, using ~\cite{Baker1999} with the substitution $r
= \twonorm{z}$, and employing integration by parts we have
\baligns
\textstyle \int_{\ball(0,1)} \twonorm{\grad\psi(z)} \dz
  &= \textstyle I_d^{-1} \int_0^1 \frac{2r}{(1-r^2)^2}\exp{\frac{-1}{1-r^2}} (d V_d
  r^{d-1})\,dr \\
  &= \textstyle \frac{d}{\theta_d} \left [-r^{d-1}\exp{\frac{-1}{1-r^2}}\bigg |_{r=0}^{r=1} + \int_0^1 (d-1)
  r^{d-2} \exp{\frac{-1}{1-r^2}}\, dr \right ] \\
  &= \textstyle \frac{d}{\theta_d} [ e^{-1}\indic{d = 1} + \indic{d \neq 1}\theta_{d-1}] = \frac{d\theta_{d-1}}{\theta_d}
\ealigns
yielding \eqnref{grad-vk-bounds}.

Finally, to prove \eqnref{hess-vk-bounds}, since $\supp{\Hess
v_{K,\delta}}\subseteq K^{2\delta}\setminus K$, we only need check for $x\in
K^{2\delta} \setminus K$. Analogous to the case $\grad v_{K,\delta}$ above, we have
\baligns
\opnorm{\Hess v_{K,\delta}(x)}
&\textstyle\le \delta^{-d-2}\int_{\ball(x,\delta)\cap K^{\delta}}
\opnorm{\Hess \psi\left ( \frac{x-y}{\delta}\right )} \dy
  \le \delta^{-2} M_1(\grad \psi) \int_{\ball(0,1)\cap K_x^{\delta}} 1 \dz
  \le \delta^{-2} M_1(\grad \psi) V_d.
\ealigns
Since
\[
\Hess \psi(x) = \textstyle I_d^{-1} \frac{\exp{-1 / (1 - \twonorm{x}^2)}}{(1
  - \twonorm{x}^2)^2} \cdot\Bigg [\frac{4}{(1 - \twonorm{x}^2)^2}xx^{\top}
  - \frac{8}{(1 - \twonorm{x}^2)} xx^{\top} - 2 I\Bigg ]
  \indic{\twonorm{x}<1},
\]
by the triangle inequality $\opnorm{\Hess \psi(x)} \le I_d^{-1}e^{-1/(1 -
r^2)}(\textfrac{4r^2}{(1 - r^2)^{4}} + \textfrac{8r^2}{(1 - r^2)^{3}} + \textfrac{2}{(1 -
r^2)^{2}}) \indic{r < 1}$ for $r = \twonorm{x}$. Hence $M_1(\grad \psi) \le 22 I_d^{-1}$ and
so $M_1(\grad v_{K,\delta}) \le \delta^{-2} \textfrac{22}{\theta_d}$ as desired.

\section{Proof of \thmref{imq-implies-weak-convergence}: IMQ KSD detects non-convergence}
\label{sec:imq-implies-weak-convergence-proof}

We first use the following theorem to upper bound the bounded Lipschitz metric $d_{\blset}(\mu, P)$ in terms of the KSD
$\langstein{\mu}{\ksteinset{k}}$.

\begin{theorem}[IMQ KSD lower bound]\label{thm:imq-bl-upper-bound}
Suppose $P \in\pset$ and $k(x,y) = (c^2 + \twonorm{x}^2)^\beta$ for $c > 0$,
and $\beta\in (-1,0)$. Choose any $\alpha \in (0,\half(\beta + 1))$ and $a >
\half c$. Then there exist an $\eps_{0} > 0$ and a constant $\mathcal{M}_P$ such that, for all $\mu$,
\balign\label{eqn:imq-bl-upper-bound}
&d_{\blset}(\mu, P) \le \inf_{\eps\in [0,\eps_{0}), \delta,\rho > 0}
  \rho \sqrt{d} (1 + M_1(b)\mathcal{M}_P) +
  \left(3 + \eps + (M_1(b)\rho\sqrt{d} + \textfrac{\delta^{-1} d
    \theta_{d-1}}{\theta_d}) \mathcal{M}_P\right )\eps +
  (2\pi)^{-d/4}\mathcal{M}_P V_d^{1/2} \times \notag \\
  &\left[\left
  (\textfrac{\mathcal{D}(a,c,\alpha,\beta)^{1/2}(\langstein{\mu}{\ksteinset{k}}
    - \zeta(a, c, \alpha, \beta))}{\alpha\kappa_0\eps}\right )^{1/\alpha} +
  2\delta \right]^{d/2}
  \sqrt{F_{IMQ}(\textfrac{12d\log 2}{\pi} (c_{\rho,\delta} + M_1(b)\mathcal{M}_P)
  \eps^{-1} )} \langstein{\mu}{\ksteinset{k}} \\
  &= O\left(\left(\log\left(\textfrac{1}{\langstein{\mu}{\ksteinset{k}}}\right)\right)^{-1/2}\right) \qtext{ as } \langstein{\mu}{\ksteinset{k}}\to 0, \label{eqn:imq-bl-upper-bound-big-o}
\ealign
for $\theta_d, \theta_{d-1}, V_d,$ and $c_{\rho,\delta}$ defined in \thmref{tightness-density-blset-upper-bound},
the function $\mathcal{D}$ defined in \eqnref{imq-coercive-norm},
the function $\zeta$ defined in \eqnref{coercive-uniform-lower-bound},
and 
\balign\label{eqn:f-mq-explicit-definition}
F_{IMQ}(t) \defeq
\textfrac{\Gamma(-\beta)}{2^{1+\beta}}\left(\textfrac{\sqrt{d}}{c}\right )^{\beta +
  d/2}\textfrac{t^{\beta+d/2}}{ K_{\beta + d/2}(c\sqrt{d} t)}
\ealign
where $K_v$ is the modified Bessel function of the third kind.
Moreover, if $\lim\sup_m \langstein{\mu_m}{\ksteinset{k}} < \infty$ then
$(\mu_m)_{m\geq 1}$ is uniformly tight.
\end{theorem}
\begin{remark}
The Stein factor $\mathcal{M}_P$ can be determined explicitly based on the proof of
\thmref{imq-bl-upper-bound} in \secref{imq-bl-upper-bound-proof} and the results of \citet{GorhamDuVoMa16}.
\end{remark}

Note that $F_{IMQ}(t)$ is finite for all $t > 0$, so fix any $\eps \in [0, \eps_0)$ and $\delta,\rho > 0$.
If $\langstein{\mu_m}{\ksteinset{k}}\to 0$, then $\lim\sup_m
d_{\blset}(\mu_m, P) \le
\rho \sqrt{d} (1 + M_1(b)\mathcal{M}_P) + (3 + \eps +
(M_1(b)\rho\sqrt{d} + \textfrac{\delta^{-1} d\theta_{d-1}}{\theta_d})
\mathcal{M}_P )\eps$.
Thus taking $\eps,\rho\to 0$ yields $d_{\blset}(\mu_m, P)\to 0$. Since $d_{\blset}(\mu_m,
P)\to 0$ only if $\mu_m\Rightarrow P$, the statement of
\thmref{imq-implies-weak-convergence} follows.

\subsection{Proof of \thmref{imq-bl-upper-bound}: IMQ KSD lower bound}
\label{sec:imq-bl-upper-bound-proof}
Fix any $\alpha\in(0,\half(\beta + 1))$ and $a > \half c$. Then there is some
$\mathring{g}\in\ksteinset{k}$ such that $\langevin{\mathring{g}}$ is
bounded below by a constant $\zeta(a, c, \alpha, \beta)$ and has a growth rate of
$\twonorm{x}^{2\alpha}$ as $\twonorm{x}\to\infty$.
Such a function exists by the following lemma, proved in
\secref{distant-dissipativity-precoercive-proof}.
\begin{lemma}[Generalized multiquadric Stein sets yield coercive functions]\label{lem:distant-dissipativity-precoercive}
Suppose $P \in\pset$ and $k(x,y) = \Phi_{c,\beta}(x-y)$ for $\Phi_{c,\beta}(x) \defeq (c^2 + \twonorm{x}^2)^\beta$, $c > 0$, and $\beta\in \reals\, \bs\, \naturals_0$.
Then, for any $\alpha \in (0, \half(\beta + 1))$ and $a > \half c$, there exists a
function $\mathring{g}\in\ksteinset{k}$ such that $\langevin{\mathring{g}}$
is bounded below by
\balign\label{eqn:coercive-uniform-lower-bound}
\zeta(a, c, \alpha, \beta) \defeq -\frac{\mathcal{D}(a,c,\alpha,\beta)^{1/2}}{2\alpha}\left [\frac{M_1(b)
    R_0^2 + \twonorm{b(0)}R_0 + d}{a^{2(1 - \alpha)}}\right ],
\ealign
where the function $\mathcal{D}$ is defined in \eqnref{imq-coercive-norm}
and $R_0 \defeq \inf\{r > 0 \,|\, \kappa(r') \ge 0, \forall r' \ge r\}$. Moreover,
$\lim\inf \twonorm{x}^{-2\alpha}\langarg{\mathring{g}}{x}\ge \frac{\alpha}{\mathcal{D}(a,
c, \alpha, \beta)^{1/2}}\kappa_0$ as $\twonorm{x}\to\infty$.
\end{lemma}

Our next lemma connects the growth rate of $\langevin{\mathring{g}}$ to the
tightness rate of a probability measure evaluated with the Stein
discrepancy.  Its proof is found in \secref{coercive-tightness-proof}.
\begin{lemma}[Coercive functions yield tightness]\label{lem:coercive-tightness}
Suppose there is a $g\in\gset$ such that $\langevin{g}$ is 
bounded below by $\zeta\in\reals$ and
$\lim\inf_{\twonorm{x}\to\infty} \twonorm{x}^{-u} \langarg{g}{x} > \eta$ for
some $\eta, u > 0$. Then for all $\eps$ sufficiently small and any probability measure $\mu$ the
tightness rate \eqnref{tightness-rate} satisfies
\baligns
R(\mu, \eps) \leq \left[\frac{1}{\eps\eta}(\langstein{\mu}{\gset} - \zeta)\right ]^{1/u}.
\ealigns
In particular, if $\lim\sup_m
\langstein{\mu_m}{\ksteinset{k}}$ is finite, $(\mu_m)_{m\ge 1}$ is
uniformly tight.
\end{lemma}

We can thus plug the tightness rate estimate of \lemref{coercive-tightness}
applied to the function $\mathring{g}$ into
\thmref{tightness-density-blset-upper-bound}. Since
$\infnorm{w}\le t$ implies $\twonorm{w}\le \sqrt{d}t$, we can use the formula for
the generalized Fourier transform of the IMQ kernel in
\eqnref{mq-kernel-gft-definition} to see $\hat{\Phi}(\omega)$ is
monotonically decreasing in $\twonorm{w}$ to establish
\eqnref{f-mq-explicit-definition}. By taking
$\eta\to\frac{\alpha}{\mathcal{D}(a, c, \alpha, \beta)^{1/2}}\kappa_0$ we
obtain \eqnref{imq-bl-upper-bound}.

To prove \eqnref{imq-bl-upper-bound-big-o}, notice that $F_{IMQ}(t) =
O(e^{(c\sqrt{d}+\lambda)t})$ as $t\to\infty$ for any $\lambda > 0$ by
\eqnref{bessel-function-inequalities}.
Hence, by choosing $\eps = \rho = ((c\sqrt{d} + 1)/\log(\textfrac{1}{\langstein{\mu}{\ksteinset{k}}}))^{1/2}$
and fixing any $\delta > 0$ we obtain the advertised decay rate as
$\langstein{\mu}{\ksteinset{k}}\to 0$.  The uniform tightness conclusion
follows from \lemref{coercive-tightness}.

\subsection{Proof of \lemref{distant-dissipativity-precoercive}: Generalized multiquadric Stein sets yield coercive functions}
\label{sec:distant-dissipativity-precoercive-proof}

By \citep[Thm.
  8.15]{Wendland2004}, $\Phi_{c,\beta}$ has a generalized Fourier transform of order
$\max(0,\lceil\beta\rceil)$ given by
\balign\label{eqn:mq-kernel-gft-definition}
\widehat{\Phi_{c,\beta}}(\omega) = \frac{2^{1+\beta}}{\Gamma(-\beta)}
\left(\frac{\twonorm{\omega}}{c}\right )^{-\beta-d/2} K_{\beta + d/2}(c\twonorm{\omega}),
\ealign
where $K_v(z)$ is the modified Bessel function of the third kind.
Furthermore, by \citep[Cor. 5.12, Lem. 5.13, Lem. 5.14]{Wendland2004}, we have the
following bounds on $K_v(z)$ for $v\in\reals, z\in (0,\infty)$:
\balign\label{eqn:bessel-function-inequalities}
K_v(z) &\ge \tau_v \frac{e^{-z}}{\sqrt{z}} \text{ for } \, z \ge 1 \text{ where }
\tau_v = \sqrt{\frac{\pi}{2}} \text{ for } |v|\ge \half \text{ and }
\tau_v = \frac{\sqrt{\pi} 3^{|v|-1/2}}{2^{|v|+1}\Gamma(|v|+1/2)} \text{ for }
|v| < \half, \\
K_v(z) &\ge e^{-1}\tau_v z^{-|v|}  \text{ for } z \le 1, \text{ (since $x\mapsto
  x^vK_{-v}(x)$ is non-increasing and $K_v = K_{-v}$)} \notag   \\
K_v(z) &\le \sqrt{\frac{2\pi}{z}} e^{-z + v^2/(2z)} \text{ for } \, z > 0, \notag \\
K_v(z) &\le 2^{|v|-1}\Gamma(|v|)z^{-|v|} \text{ for } \, v\neq 0, z > 0. \notag
\ealign

Now fix any $a >c/2$ and $\alpha \in(0,\half(\beta + 1))$, and consider the functions $g_j(x) = \grad_{x_j} \Phi_{a,
  \alpha}(x) = 2\alpha x_j (a^2 + \twonorm{x}^2)^{\alpha - 1}$. 
We will show that $g= (g_1, \dots,g_d)\in\kset_k^d$. Note that $\hat{g_j}(\omega) =
(i\omega_j)\widehat{\Phi_{a,\alpha}}(\omega)$. Using \citep[Thm.
  10.21]{Wendland2004}, we know $\knorm{g_j} =
\norm{\hat{g_j}/\sqrt{\widehat{\Phi_{c,\beta}}}}_{L^2}$, and thus
$\kdnorm{g} =
\norm{\hat{g}/\sqrt{\widehat{\Phi_{c,\beta}}}}_{L^2}$. Hence
\baligns
\kdnorm{g}^2
  &= \sum_{j=1}^d \int_{\reals^d}
    \hat{g_j}(\omega)\overline{\hat{g_j}(\omega)}/\widehat{\Phi_{c,\beta}}(\omega)\,d\omega \\
  &= \sum_{j=1}^d \int_{\reals^d} \frac{2^{2(1 + \alpha)} / \Gamma(-\alpha)^2}{2^{1 +
    \beta} / \Gamma(-\beta)} \frac{a^{2\alpha + d}}{c^{\beta + d/2}}
  \omega_j^2 \twonorm{\omega}^{\beta - 2\alpha - d/2} \frac{K_{\alpha +
      d/2}(a\twonorm{\omega})^2}{K_{\beta + d/2}(c\twonorm{\omega})} \, d\omega \\
  &= c_0 \int_{\reals^d} \twonorm{\omega}^{\beta - 2\alpha - d/2 + 2} \frac{K_{\alpha +
      d/2}(a\twonorm{\omega})^2}{K_{\beta + d/2}(c\twonorm{\omega})} \,
      d\omega,
\ealigns
where $c_0 = \frac{2^{2(1 + \alpha)} / \Gamma(-\alpha)^2}{2^{1 + \beta} /
  \Gamma(-\beta)} \frac{a^{2\alpha + d}}{c^{\beta + d/2}}$.
We can split the integral above into two, with the first integrating over
$\ball(0,1)$ and the second integrating over $\ball(0,1)^c =
\reals^d\setminus\ball(0,1)$. Thus using the inequalities
from \eqnref{bessel-function-inequalities} with $v_0 \defeq \beta + d/2$,
we have
\baligns
\int_{\ball(0,1)} \twonorm{\omega}^{\beta - 2\alpha - d/2 + 2} \frac{K_{\alpha +
  d/2}(a\twonorm{\omega})^2}{K_{\beta + d/2}(c\twonorm{\omega})} \,
  d\omega
  &\le \int_{\ball(0,1)} \twonorm{\omega}^{\beta - 2\alpha - d/2 + 2}
  \frac{2^{2\alpha + d-2}\Gamma(\alpha+d/2)^2(a\twonorm{w})^{-2\alpha-d}}
       {e^{-1}\tau_{v_0}\cdot \twonorm{c\omega}^{-\beta-d/2}} \,d\omega \\
  &= 2^{2\alpha + d-2} \Gamma(\alpha+d/2)^2\frac{e}{\tau_{v_0}}
  \frac{c^{\beta+d/2}}{a^{2\alpha+d}} \int_{\ball(0,1)} \twonorm{\omega}^{2\beta -
    4\alpha - d + 2} e^{c\twonorm{\omega}}\,d\omega \\
  &= d\, V_d\, 2^{2\alpha + d- 2} \Gamma(\alpha+d/2)^2\frac{e}{\tau_{v_0}}
  \frac{c^{\beta+d/2}}{a^{2\alpha+d}}
  \int_0^1 r^{2\beta - 4\alpha + 1} e^{c r}\,dr,
\ealigns
where $V_d$ is the volume of the unit ball in $d$-dimensions and in the last
step we used the substitution $r = \twonorm{\omega}$ \cite{Baker1999}. Since
$\alpha < \half(\beta+1)$ and the function $r\mapsto r^t$ is integrable around the
origin when $t > -1$, we can bound the integral above by
\baligns
\int_0^1 r^{2\beta - 4\alpha + 1} e^{c r}\,dr
  \le e^c \int_0^1 r^{2\beta - 4\alpha + 1}\,dr
  = \frac{e^c}{2\beta - 4\alpha + 2}.
\ealigns

We can apply the technique to the other integral, yielding
\baligns
\int_{\ball(0,1)^c} \twonorm{\omega}^{\beta - 2\alpha - d/2+2} \frac{K_{\alpha +
  d/2}(a\twonorm{\omega})^2}{K_{\beta + d/2}(c\twonorm{\omega})} \,
  d\omega
  &\le \int_{\ball(0,1)^c} \twonorm{\omega}^{\beta - 2\alpha - d/2 + 2}
  \frac{2\pi / (a\twonorm{\omega}) \cdot e^{-2a\twonorm{\omega} + (\alpha + d/2)^2
      / (a\twonorm{\omega})}}{\tau_{v_0} e^{-c\twonorm{\omega}}/\sqrt{c\twonorm{\omega}}} \, d\omega \\
  &\le \frac{2\pi\sqrt{c}}{a\tau_{v_0}}\int_{\ball(0,1)^c} \twonorm{\omega}^{\beta
    - 2\alpha - d/2 + 3/2} e^{(c - 2a)\twonorm{\omega} + (\alpha +
    d/2)^2/(a\twonorm{\omega})} \, d\omega \\
  &= d\, V_d\, \frac{2\pi\sqrt{c}}{a\tau_{v_0}}\int_{1}^{\infty} r^{\beta
    - 2\alpha + d/2 + 1/2} e^{(c - 2a)r + (\alpha + d/2)^2/(ar)} \, dr
\ealigns
Since $c - 2a < 0$, we can upper bound the last integral above by the quantity
\baligns
\int_{1}^{\infty} r^{\beta - 2\alpha + d/2 + 1/2} &e^{(c - 2a)r + (\alpha +
d/2)^2/(ar)} \, dr \\
  &\le e^{(c - 2a) + (\alpha + d/2)^2/a} \int_{1}^{\infty} r^{\beta -
  2\alpha + d/2 + 1/2} e^{(c - 2a)r} \, dr \\
  &= e^{(c - 2a) + (\alpha + d/2)^2/a} (2a - c)^{-\beta + 2\alpha - d/2 -
  3/2} \Gamma(\beta - 2\alpha + d/2 + 3/2, 2a - c),
\ealigns
where $\Gamma(s,x) \defeq \int_x^{\infty} t^{s-1}e^{-t}\dt$ is the upper
incomplete gamma function.  Hence, the function $g$ belongs to $\kset_k^d$
with norm upper bounded by $\mathcal{D}(a, b, \alpha, \beta)^{1/2}$ where
\balign\label{eqn:imq-coercive-norm}
\mathcal{D}(a, c, \alpha, \beta) &\defeq d\, V_d\, 2^{1 + 2\alpha - \beta}
  \frac{a^{2\alpha + d} \Gamma(-\beta)}{c^{\beta + d/2} \Gamma(-\alpha)^2} \Bigg (
\frac{2^{2\alpha + d- 2} \Gamma(\alpha+d/2)^2 e^{c+1} c^{\beta + d/2}}
     {\tau_{v_0} (2\beta - 4\alpha + 2)a^{2\alpha + d}} + \notag \\
&\frac{2\pi\sqrt{c}}{a\tau_{v_0}} e^{(c - 2a) + (\alpha + d/2)^2/a}
(2a - c)^{-\beta + 2\alpha - d/2 - 3/2} \Gamma(\beta - 2\alpha + d/2 + 3/2,
2a - c)
\Bigg ).
\ealign

Now define
$\mathring{g} = -\mathcal{D}(a, c, \alpha, \beta)^{-1/2} g$ so that
$\mathring{g}\in\ksteinset{k}$. We will lower bound the growth rate of
$\langevin{\mathring{g}}$ and also construct a uniform lower bound. Note
\balign\label{eqn:precoercive-full-form}
\frac{\mathcal{D}(a, c, \alpha, \beta)^{1/2}}{2\alpha} \langarg{\mathring{g}}{x} &= -\frac{\inner{b(x)}{x}}{(a^2 +
  \twonorm{x}^2)^{1-\alpha}} - \frac{d}{(a^2 +
  \twonorm{x}^2)^{1-\alpha}} + \frac{2(1-\alpha)\twonorm{x}^2}{(a^2 +
  \twonorm{x}^2)^{2-\alpha}}.
\ealign
The latter two terms are both uniformly bounded in $x$. By the distant
dissipativity assumption, there is some $\kappa > 0$
such that
$\lim\sup_{\twonorm{x}\to\infty} \frac{1}{\twonorm{x}^2}\inner{b(x)}{x}\le
-\half\kappa$. Thus the first term
of \eqnref{precoercive-full-form} grows at least at the rate
$\half\kappa\twonorm{x}^{2\alpha}$. This assures
$\lim\inf \twonorm{x}^{-2\alpha} \langarg{\mathring{g}}{x} \ge \frac{\alpha}{\mathcal{D}(a,
c, \alpha, \beta)^{1/2}}\kappa$ as $\twonorm{x}\to\infty$.

Moreover, because $b$ is Lipschitz, we have
\baligns
|\inner{b(x)}{x}|
  \le |\inner{b(x) - b(0)}{x - 0}| + |\inner{b(0)}{x}|
  \le M_1(b)\twonorm{x}^2 + \norm{b(0)}\twonorm{x},
\ealigns
Hence for any $x\in\ball(0, R_0)$, we must have $-\inner{b(x)}{x}\ge
-M_1(b)R_0^2 - \twonorm{b(0)}R_0$. By choice of $R_0$,
for all $x\notin\ball(0, R_0)$, the distant dissipativity
assumption implies $-\inner{b(x)}{x}\ge 0$. Hence applying this to
\eqnref{precoercive-full-form} shows that $\langevin{\mathring{g}}$ is
uniformly lower bounded by $\zeta(a, c, \alpha, \beta)$.

\subsection{Proof of \lemref{coercive-tightness}: Coercive functions yield tightness}
\label{sec:coercive-tightness-proof}

Pick $g\in\ksteinset{k}$ such that
$\lim\inf_{\twonorm{x}\to\infty} \twonorm{x}^{-u}\langarg{g}{x} > \eta$ and
$\inf_{x\in\reals^d} \langarg{g}{x}\ge \zeta$.
Let us define $\gamma(r) \defeq \inf \{\langarg{g}{x} - \zeta \,|\,
\twonorm{x}\ge r\}\geq 0$ for all $r > 0$.
Thus for sufficiently large $r$, we have $\gamma(r) \ge \eta r^u$.
Then, for any measure $\mu$ by Markov's inequality, 
\baligns
\mu(\twonorm{\QVAR} \ge r)
  \le \frac{\Esubarg{\mu}{\gamma(\twonorm{\QVAR})}}{\gamma(r)}
  \le \frac{\Esubarg{\mu}{\langarg{g}{\QVAR} - \zeta}}{\gamma(r)}.
\ealigns
Thus we see that $\mu(\twonorm{\QVAR} \ge r_{\eps})
\le \eps$ whenever $\eps \ge (\langstein{\mu}{\ksteinset{k}} - \zeta)/\gamma(r_{\eps})$.
This implies that for sufficiently small $\eps$, if
\baligns
r_{\eps} \ge \left[\frac{1}{\eta \eps} (\langstein{\mu}{\ksteinset{k}} -
  \zeta)\right ]^{1/u},
\ealigns
we must have $\mu(\twonorm{\QVAR}\ge r_{\eps})\le \eps$. Hence whenever
$\lim\sup_m \langstein{\mu_m}{\ksteinset{k}}$ is bounded, we must have
$(\mu_m)_{m\ge 1}$ is uniformly tight as $\lim\sup_m R(\mu_m, \eps)$ is
finite.

\section{Proof of \propref{ksd-upper-bound}: KSD detects convergence}\label{sec:ksd-upper-bound-proof}

We will first state and prove a useful lemma.

\begin{lemma}[Stein output upper bound]\label{lem:kset-function-upper-bound}
Let $Z\sim P$ and $X\sim\mu$. If the score function $b=\grad \log p$ is Lipschitz
with $\Esub{P}[\twonorm{b(Z)}^2] < \infty$, then, for any $g:\reals^d \to \reals^d$
with $\max(M_0(g), M_1(g), M_2(g)) < \infty$,
\baligns%
|\Esubarg{\mu}{\langarg{g}{\QVAR}}|
    \le
   (M_0(g)M_1(b) + M_2(g)d)\twowass(\mu,P) +
    \sqrt{2M_0(g)\,M_1(g)\,
     \Esub{P}[\twonorm{b(\PVAR)}^2]\,
     \twowass(\mu,P)},
\ealigns
where the Wasserstein distance $\twowass(\mu,P) = \inf_{\QVAR\sim \mu, \PVAR \sim P} \Earg{\twonorm{\QVAR - \PVAR}}$.
\end{lemma}
\begin{proof}
By Jensen's inequality, we have
$\Esubarg{P}{\twonorm{b(Z)}} \le \sqrt{\Esub{P}[\twonorm{b(Z)}^2]} <
\infty$, which implies that $\Esubarg{P}{\langarg{g}{\PVAR}} = 0$
\citep[Prop. 1]{GorhamMa15}.
Thus, using the triangle inequality, Jensen's inequality, and the Fenchel-Young inequality for dual norms, 
\baligns
|\Esubarg{\mu}{\langarg{g}{\QVAR}}| &= |\Earg{\langarg{g}{\PVAR} - \langarg{g}{\QVAR}}| \\
  &= |\Earg{\inner{b(\PVAR)}{g(\PVAR) - g(\QVAR)} + \inner{b(\PVAR) - b(\QVAR)}{g(\QVAR)} + \inner{I}{\grad g(\PVAR) - \grad g(\QVAR)}}| \\
  &\le \Earg{|\inner{b(\PVAR)}{g(\PVAR) -
  g(\QVAR)}|} + (M_0(g)M_1(b) + M_2(g) d)\Earg{\twonorm{\QVAR-\PVAR}},
\ealigns
 To handle the other term above, notice that by Cauchy-Schwarz and
the fact that $\min(a,b)\le \sqrt{ab}$ for $a,b \ge 0$,
\baligns
\Earg{|\inner{b(\PVAR)}{g(\PVAR) - g(\QVAR)}|}
  &\le \Earg{\min\left (2M_0(g),
   M_1(g)\twonorm{\QVAR-\PVAR}\right )\twonorm{b(\PVAR)}} \\
 &\le (2 M_0(g) M_1(g))^{1/2}
   \Earg{\twonorm{\QVAR-\PVAR}^{1/2} \twonorm{b(\PVAR)}} \\
 &\le \sqrt{2 M_0(g) M_1(g)
   \Earg{\twonorm{\QVAR-\PVAR}}\,
   \Esub{P}[\twonorm{b(\PVAR)}^2]}.
\ealigns
The stated inequality now follows by taking the infimum of these bounds over all joint distributions $(X, Z)$ with $X\sim \mu$ and
$Z\sim P$.
\end{proof}

Now we are ready to prove \propref{ksd-upper-bound}.  
In the statement below, let us use $\alpha \in \naturals^d$ as a multi-index
for the differentiation operator $D^{\alpha}$, that is, for a differentiable
function $f:\reals^d\to\reals$ we have for all $x\in\reals^d$,
\baligns
D^{\alpha}f(x) \defeq \frac{d^{|\alpha|}}{(dx_1)^{\alpha_1}\dots (dx_d)^{\alpha_d}}
f(x)
\ealigns
where $|\alpha| = \sum_{j=1}^d \alpha_j$.
Pick any
$g\in\ksteinset{k}$, and choose any multi-index $\alpha \in \naturals^d$ such that
$|\alpha|\le 2$. Then by Cauchy-Schwarz and \citep[Lem.
  4.34]{Christmann2008}, we have
\baligns
\sup_{x\in\reals^d} |D^{\alpha} g_j(x)|
  = \sup_{x\in\reals^d} |D^{\alpha} \inner{g_j}{k(x,\cdot)}_{\kset_k}|
  \le \sup_{x\in\reals^d} \knorm{g_j}\, \knorm{D^{\alpha}
  k(x,\cdot)}
  &= \knorm{g_j} \sup_{x\in\reals^d}\, (D_x^{\alpha}D_y^{\alpha}
    k(x,x))^{1/2}.
\ealigns
Since $\sum_{j=1}^d \knorm{g_j}^2 \leq 1$ for all $g\in \ksteinset{k}$ and
$D_x^{\alpha}D_y^{\alpha} k(x,x)$ is uniformly bounded in $x$ for all
$|\alpha| \le 2$, the elements of the vector $g(x)$, matrix $\grad
g(x)$, and tensor $\Hess g(x)$ are uniformly bounded in $x\in\reals^d$ and $g\in\ksteinset{k}$. 
Hence, for some $\lambda_k$,
$\sup_{g\in\ksteinset{k}}\max(M_0(g), M_1(g), M_2(g))\leq \lambda_k < \infty$, so the advertised result follows from
\lemref{kset-function-upper-bound} as
\[
\langstein{\mu}{\ksteinset{k}} 
	\leq \lambda_k \left( (M_1(b) + d)\twowass(\mu,P) +
    \sqrt{2
     \Esub{P}[\twonorm{b(\PVAR)}^2]\,
     \twowass(\mu,P)}\right).
\]

\section{Proof of \thmref{bounded-score-function}: KSD fails for bounded scores}\label{sec:bounded-score-function-proof}

Fix some $n\ge 1$, and let $Q_n = \frac{1}{n}\sum_{i=1}^n \delta_{x_i}$ where $x_i
\defeq i n e_1 \in \reals^d$ for $i\in\{1,\dots, n\}$. This implies
$\twonorm{x_i - x_{i'}} \ge n$ for all $i\neq i'$. We will show
that when $M_0(b)$ is finite, $\langstein{Q_n}{\ksteinset{k}}\to 0$ as
$n\to\infty$.

We can express $k_0(x,y) \defeq \sum_{j=1}^d k_0^j(x,y)$ as
\baligns k_0(x,y)
  &= \inner{b(x)}{b(y)}k(x,y)
  + \inner{b(x)}{\grad_y k(x,y)}
  + \inner{b(y)}{\grad_x k(x,y)}
  + \inner{\grad_x}{\grad_y k(x,y)}.
\ealigns
From \propref{kernel-stein-discrepancy-formula}, we have
\balign\label{eqn:ksd-bounded-score-function-decomp}
\langstein{Q_n}{\ksteinset{k}}^2
  = \frac{1}{n^2}\sum_{i,i'=1}^n k_0(x_i, x_{i'})
  = \frac{1}{n^2}\sum_{i=1}^n k_0(x_i, x_i) + \frac{1}{n^2}\sum_{i\neq i'} k_0(x_i, x_{i'}).
\ealign
Let $\gamma$ be the kernel decay rate defined in the statement of
\thmref{counterexample-theorem}.
Then as $k\in C^{(1,1)}_0$, we must have $\gamma(0) < \infty$ and
$\lim_{r\to\infty} \gamma(r) = 0$. By the triangle inequality
\baligns
\lim_{n\to\infty} \left |\frac{1}{n^2}\sum_{i=1}^n k_0(x_i,x_i)\right |
  \le \lim_{n\to\infty} \frac{1}{n^2}\sum_{i=1}^n |k_0(x_i,x_i)|
  \le \lim_{n\to\infty} \frac{\gamma(0)}{n} (M_0(b) + 1)^2
  = 0.
\ealigns
We now handle the second term of \eqnref{ksd-bounded-score-function-decomp}.
By repeated use of Cauchy-Schwarz we have
\baligns
|k_0(x_i, x_{i'})|
  &\le |\inner{b(x_i)}{b(x_{i'})} k(x_i, x_{i'})|
  + |\inner{b(x_i)}{\grad_y k(x_{i}, x_{i'})}|
  + |\inner{b(x_{i'})}{\grad_x k(x_i, x_{i'})}|
  + |\inner{\grad_x}{\grad_y k(x_i, x_{i'})}| \\
  &\le \twonorm{b(x_i)}\twonorm{b(x_{i'})} |k(x_i, x_{i'})|
  + \twonorm{b(x_{i})}\twonorm{\grad_y k(x_{i}, x_{i'})}
  + \twonorm{b(x_{i'})}\twonorm{\grad_x k(x_i, x_{i'})} \\
  &\qquad + |\inner{\grad_x}{\grad_y k(x_i, x_{i'})}| \\
  &\le \gamma(n) (M_0(b) + 1)^2.
\ealigns
By assumption, $\gamma(r) \to 0$ as $r \to \infty$.
Furthermore, since the second term of
\eqnref{ksd-bounded-score-function-decomp} is upper bounded by the average
of the terms $k_0(x_i, x_i')$ for $i\neq i'$, we have
$\langstein{Q_n}{\ksteinset{k}}\to 0$ as $n\to\infty$.
However, $(Q_n)_{n\ge 1}$ is not uniformly tight and hence does not converge
to the probability measure $P$.

 \end{document}